\documentclass[12pt]{article}

\usepackage{newtxtext,newtxmath}

\usepackage{graphicx}
\usepackage{hyperref}

\usepackage[letterpaper,margin=1in]{geometry}
\usepackage{xcolor}

\renewenvironment{abstract}
	{\quotation}
	{\endquotation}

\date{}

\makeatletter
\renewcommand{\fnum@figure}{\textbf{Figure \thefigure}}
\renewcommand{\fnum@table}{\textbf{Table \thetable}}
\makeatother

\usepackage{scicite}

\usepackage{url}

\def\scititle{
	Combining digital data streams and epidemic networks for real time outbreak detection
}
\title{\bfseries \boldmath \scititle}

\author{
	Ruiqi~Lyu$^{1\ast}$,
	Alistair~Turcan$^{1}$,
	Bryan~Wilder$^{1}$\and
	\small$^{1}$School of Computer Science, Carnegie Mellon University, Pittsburgh, PA, USA.\and
	\small$^\ast$Corresponding author. Email: ruiqil@cs.cmu.edu
}

\begin{document} 

\maketitle

\begin{abstract} \bfseries \boldmath
    
    Responding to disease outbreaks requires close surveillance of their trajectories, but outbreak detection is hindered by the high noise in epidemic time series. Aggregating information across data sources has shown great denoising ability in other fields, but remains underexplored in epidemiology. Here, we present LRTrend, an interpretable machine learning framework to identify outbreaks in real time. LRTrend effectively aggregates diverse health and behavioral data streams within one region and learns disease-specific epidemic networks to aggregate information across regions. We reveal diverse epidemic clusters and connections across the United States that are not well explained by commonly used human mobility networks and may be informative for future public health coordination. We apply LRTrend to 2 years of COVID-19 data in 305 hospital referral regions and frequently detect regional Delta and Omicron waves within 2 weeks of the outbreak's start, when case counts are a small fraction of the wave’s resulting peak.

\end{abstract}

\section*{Introduction}
\noindent

Many public health interventions rely on timely warning of an emerging upswing in disease spread to arrange supply chains, impose non-pharmaceutical interventions, or prepare health systems \cite{chinazzi2020effect}. Health data time-series are crucial in these decision making processes when responding to pandemics \cite{viboud2016generalized}. However, most metrics for tracking disease prevalence such as confirmed cases, hospitalizations, and deaths are subject to availability issues and uncertainties in the data. Day-of-week effects and data collection processes leave it difficult to simply observe a stream's recent data points and conclude there is an increasing trend. For example, case counts often represent increased testing availability rather than disease prevalence, and these cases may trail the infection start time by days or weeks \cite{kaashoek2020covid}. A timely method, robust to these data artifacts, would greatly aid in implementing public health interventions.

Current assessment of disease activity is largely reliant on direct measures, such as confirmed case and death counts, which typically come well after infection. However, an enormous amount of data is collected every day and used to train machine learning models to predict human behavior in ways not apparent through analyzing a single data stream \cite{liu2020real, lu2021internet, lampos2021tracking}. Internet data such as Google symptom searches may be an early outbreak indicator and are readily available \cite{bavadekar2020google}. The Delphi US COVID-19 Trends and Impact Survey (CTIS) was conducted on Facebook to gather several proxy metrics of COVID-19 prevalence among communities \cite{salomon2021us} without explicitly relying on healthcare service availability. Many studies have explored taking advantage of these digital data streams to track population level disease changes \cite{yang2015argo, mcgough2017zika, santillana2015combining, dugas2013googleflu, lee2017socialmedia, aiken2020realtime, lu2018boston, lu2019nowcasting, stolerman2023using}. Where sensitive, gold standard data streams may not be readily available, these indirect streams can be. They can further harbor orthogonal information, considering data may be collected before testing ever occurs \cite{salathe2012digital, santillana2016perspectives}.

Much of previous work has focused on forecasting case counts or deaths, such as the CDC's COVID-19 Forecasting Hub Consortium's models \cite{cramer2022united}. This is distinct from trend detection, which aims to identify if a region is currently in the early stages of an upswing. Detecting sharp increases in trends remains difficult; as noted by Cramer \textit{et al} \cite{cramer2022evaluation}, forecasting models "should not be relied upon for decisions about the possibility or timing of rapid changes in trends". Many statistical methods do exist for detecting swiftly increasing trends. The Mann-Kendall test assesses trend significance, but is not suited for the weekly periodicity of public health data \cite{yue2002power}. Other denoising and filtering techniques, including discrete wavelet transformation \cite{nason1995stationary} and L1 or L2 trend filtering \cite{kim2009ell_1} penalize the variability of fitted curves to extract trends. A key limitation of these methods is the lack of a controllable bandwidth to specify the desired timescale for detecting trends, limiting their interpretability as to when the outbreak is occurring. Specific to epidemiology, Stolerman \textit{et al} \cite{stolerman2023using} model data streams with linear regression, binarizing and summing growth rates, later thresholding this sum to create a multivariate alarm. While effectively taking advantage of multiple streams within a region, repeated thresholding and binarization will inherently come at the cost of resolution. Leveraging information across regions is less explored, with what work does exist encountering difficulties in using cross-region information effectively. For instance, it has recently been discovered in influenza forecasting that aggregating neighboring regions' information is no better than incorporating the national average. \cite{THIVIERGE2025100820}

Here, we introduce LRTrend (Local Regression to identify Trends), a simple, interpretable machine learning framework to detect sharp increases in trends in near real-time. LRTrend employs local regression to smooth time-series and estimate the growth rate within a specified window size at high resolution. Local regression offers several advantages. First, it does not require thresholding or binarization, offering smooth and interpretable estimates of trend upswings. 
Second, it is irrespective to the data modality and fits with any distributional assumptions.
Furthermore, LRTrend naturally extends to multi-stream aggregation methods, enabling multiple powerful denoising methods to aid in low signal-to-noise circumstances. 
First, information is aggregated across multiple data streams within one region, such as case counts, deaths, or search trends, to denoise within-region streams. Second, information is aggregated across multiple geographic regions by learning a nation-wide epidemic network structure from historical data to denoise across regions and produce more accurate estimates of growth rates when a single region may have low signal.





We conduct experiments on 12 epidemiological data streams within 305 Hospital Referral Regions (HRRs), representing geographic units where patients are likely to be referred to the same hospital to receive tertiary medical care, often crossing political boundaries like state or county.
We construct a ground truth by applying trend filters to 3 gold standard data streams with full retrospective information. LRTrend can almost completely recapitulate ground truth trends in real time with less than two week's delay and when case counts are a small fraction of the wave's peak. We show that while a single data stream may be underpowered when it is not directly disease-relevant, incorporating multiple low-power data streams into a single multivariate model can achieve comparable power to that of a higher quality signal. We explore the potential of denoising across US-wide networks relevant to disease spread and show that our learned network structure strongly improves detection capability, but, surprisingly, is not well explained by mobility, commuting flows, or distance, and aggregating across these commonly used networks actually blur data streams and reduce power. Interpreting LRTrend's learned epidemic networks yields complex epidemic clusters and connections which may be informative for future disease response and coordination.

\section*{Results}

\subsection*{Method overview}

\begin{figure} 
	\centering
	\includegraphics[width=1.0\textwidth]{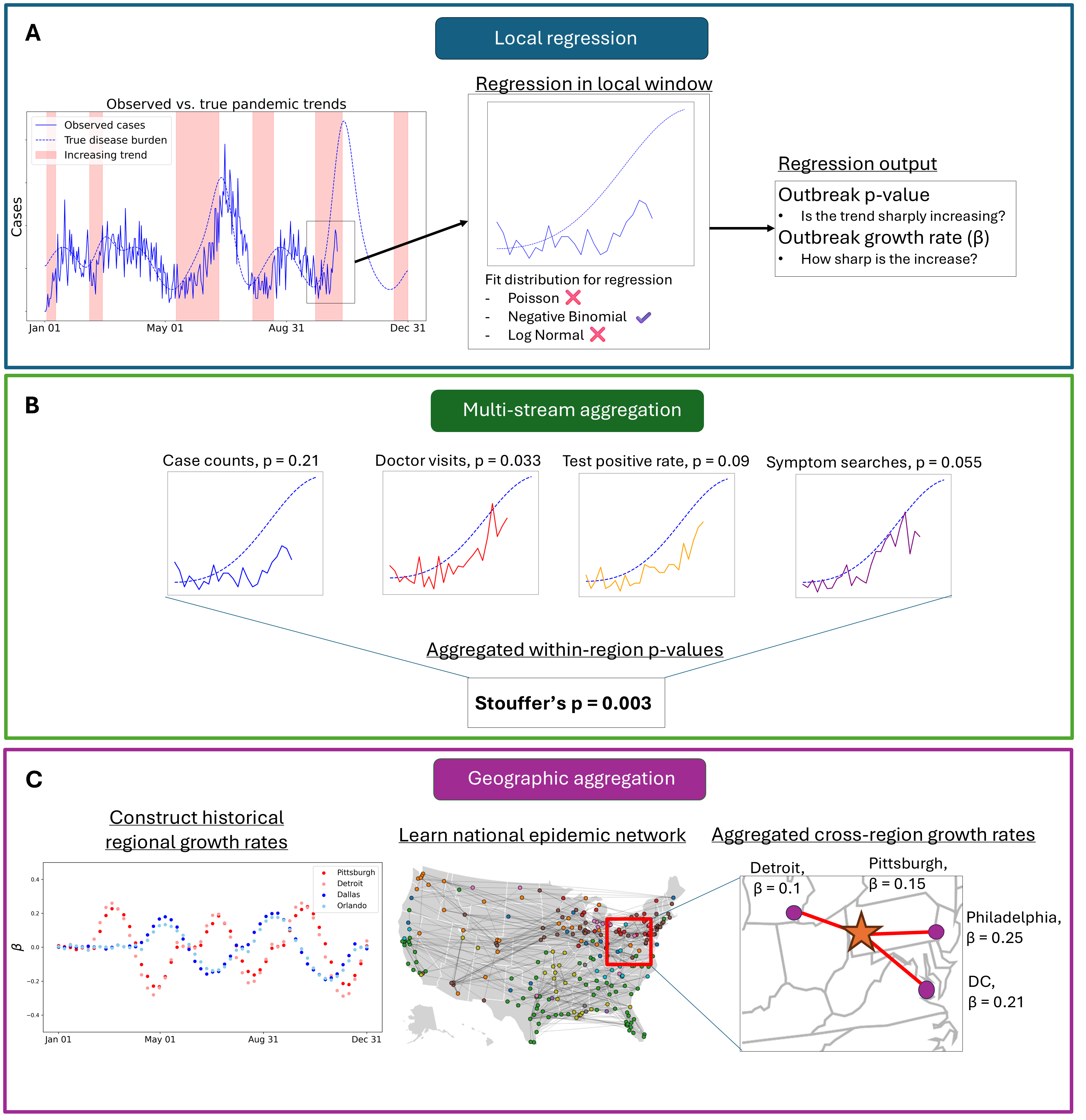} 

	\caption{\textbf{Overview of LRTrend.} (\textbf{A}) Univariate local regression pipeline. Example real-time case data (blue line) versus ground truth disease prevalence (dashed blue line) and increasing trends (shaded red) is shown first. LRTrend operates on the recent window (black box) to output $p$-values and growth rates for that window. (\textbf{B}) Multivariate local regression pipeline. 4 example data streams are shown for a given window, each with their own local regression results. (\textbf{C}) Geographic aggregation pipeline. Historical growth rates are defined from previous windows for a given data stream. Pittsburgh is the focal region, others are learned neighboring regions.}
	\label{fig:model_overview} 
\end{figure}

LRTrend takes as input one or more real-time epidemiological data streams and assesses if the growth rate is exceedingly high within a specified window, signaling a sharp increase. LRTrend consists of 3 steps (Fig. \ref{fig:model_overview}). First, within a recent window, local regression is performed on counts or rates to obtain estimates of the growth rate $\beta$ and a empirical $p$-value for if the trend is sharply increasing. This regression works in conjunction with 3 distributions commonly assumed in epidemiological data; Poisson, Log Normal, and Negative Binomial, as different data streams may be modeled better with different assumptions. Second, if multiple data streams are available, LRTrend can model each individually and subsequently combine their empirical $p$-values into a single, more powerful test using Stouffer's method of $p$-value combination. Third, LRTrend learns a national epidemic network structure by constructing informative regional features from historical time series. Epidemiologically similar regions, which may be geographically distant, can be averaged across to denoise growth rates and increase power. See Materials and Methods for full details.



Given the lack of ground truth of what constitutes a "sharp" increase, we use retrospective methods on gold standard data streams to construct a set of high-confidence trends, and aim to identify these trends using only real-time information from diverse data streams. We develop a retrospective smoother that builds upon the Delphi Epidata API and Rumack \textit{et al} \cite{rumack2023modeling}. We extend these to include Log Normal smoothing in addition to Poisson in the observation of many data streams exhibiting overdispersion \cite{xekalaki2016under}. Furthermore, we enable these smoothers to use multiple data streams as input, each with their own distributional assumptions, to avoid constructing ground truth using only a single data stream. We run this retrospective method across a variety of hyperparameter settings, and take the consensus regions (all growth rates positive) as the ground truth trends to identify. Full retrospective method details are available in the Materials and Methods.

We analyze 12 COVID-19 daily data streams (Materials and Methods), consisting of directly disease-relevant signals such as case counts and doctor visits, as well as internet based proxy metrics such as Google Search trends and Facebook surveys, in conjunction with 305 hospital referral regions (HRRs) in the United States between 2021-01-07 and 2022-06-21. We define ground truth using our retrospective method on the high quality, gold standard streams COVID-19 admissions, JHU COVID-19 case counts, and Change Healthcare COVID-19 claims, and aim to recapitulate this ground truth with the 12 real-time data streams. 
We measure model performance using power, as the percent of ground-truth trends identified, as well as delay, the days after an outbreak when it is discovered (default 60 days if a trend goes undetected), both measured at $5\%$ false positive rate (FPR).

\subsection*{Applying LRTrend with univariate data streams}

We construct high-confidence trends for all HRRs with sufficient signal in gold standard streams following our retrospective model, resulting in a total of 488 increasing trends in 152 HRRs between 2021-01-07 and 2022-06-21 (average 3.2 trends spanning 51 days each). We apply univariate LRTrend to each of the 12 data streams in each HRR at fixed FPR 5\% to annotate regions of significant increase, interpreting the start of a trend as an alarm that the disease has begun to rapidly spread.

\begin{figure} 
	\centering
	\includegraphics[width=1.0\textwidth]{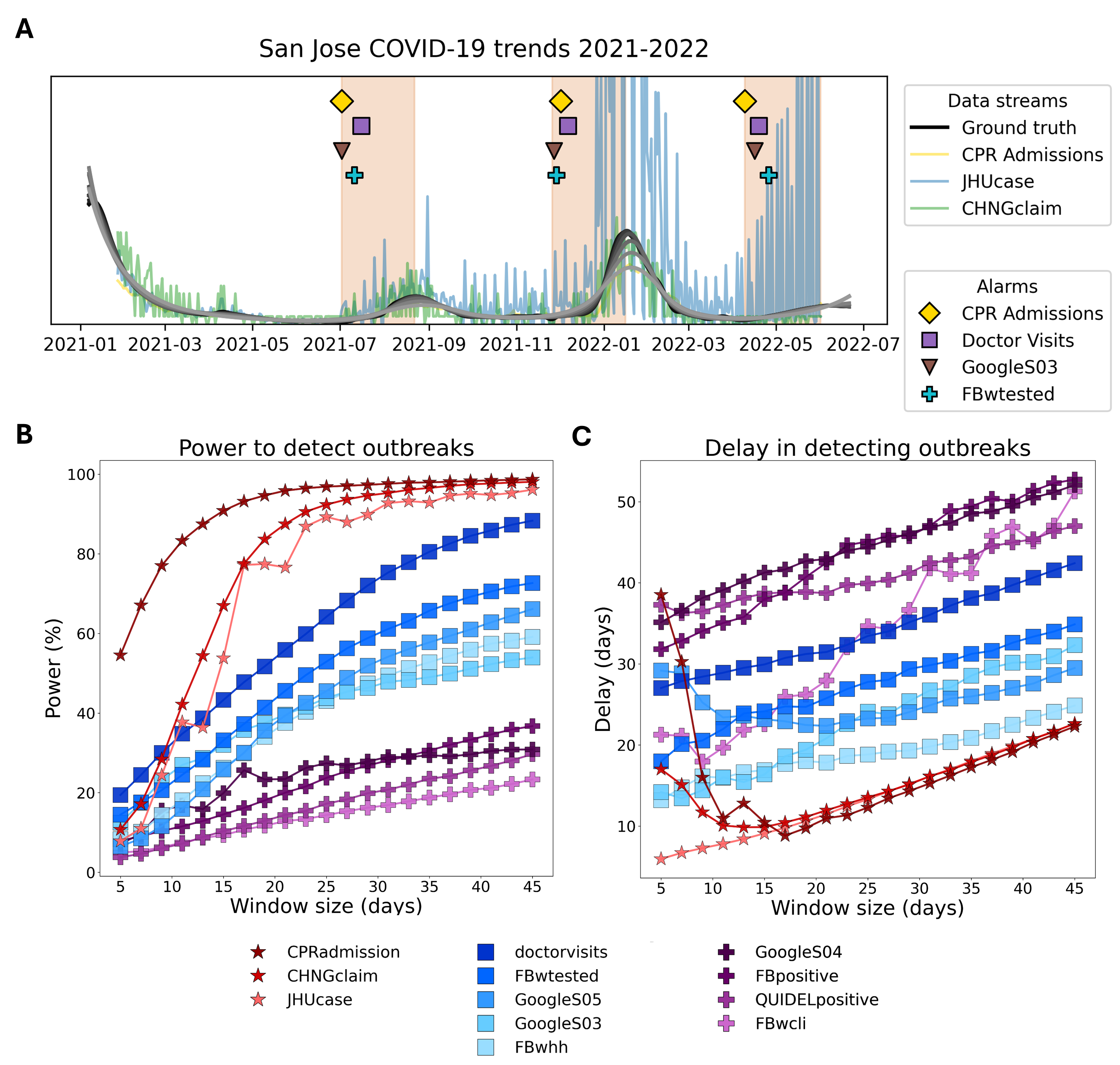} 

	\caption{\textbf{Assessing individual stream performance.} (\textbf{A}) Raw COVID-19 data streams CPR admissions, JHU cases, and Change Healthcare Claims. LRTrend's retrospective ground truth is indicated with a solid gray/black line, colored differently for different penalty values. Consensus outbreak regions are annotated in red shaded areas. Alarms are annotated from applying LRTrend with each data stream. (\textbf{B}) LRTrend's power in detecting outbreaks in conjunction with each data stream versus window size used for detection. GT streams are colored red, Medium streams colored blue, and Weak streams colored purple. (\textbf{C}) LRTrend's delay in detecting outbreaks in conjunction with each data stream versus window size used for detection. GT streams are colored red, Medium streams colored blue, and Weak streams colored purple. }
	\label{fig:univariate} 
\end{figure}

First, we visualize our task and validate our retrospective ground truth in Fig. \ref{fig:univariate}A. In the San Jose region from January 1st 2021 to June 21st 2022, there were 3 significant trends starting in July 2021, December 2021, and April 2022, consistent across parameter settings for LRTrend's retrospective smoother. Visually, these align with increasing trends for gold standard COVID data streams. In the literature, these 3 waves align with the Delta wave originating around Spring 2021 \cite{truelove2022projected}, the Omicron BA.1 wave originating around Winter 2022 \cite{lewnard2022clinical}, and the Omicron BA.2 wave originating around Spring 2022 \cite{ma2023genomic}, respectively. We plot these trends for 8 more HRRs across 5 states (Supplementary Fig. \ref{fig:time_streams}), observing strong consistency between LRTrend's constructed ground truth and visually increasing trends, as well as known COVID-19 waves like Omicron and Delta. Next, we examine how LRTrend might indicate an alarm for this region. Applying LRTrend to 4 representative real-time data streams with a fixed window size of 21 days, we see that LRTrend often triggers alarms just after the start of an outbreak when case counts are still small, regardless of whether the stream used for detection is part of the ground truth.

We next explore the performance of individual streams systematically across all HRRs in Figs. \ref{fig:univariate}B and C. We first observe that the 3 streams comprising the ground truth can exhibit strong power to detect outbreaks with only real-time data, capable of detecting 80-95\% of true trends with less than 2 weeks delay when using a window size of 3 weeks. This is likely due to both being used to defined ground truth, and the strong disease proximity of these signals. Alternate data streams such as \textit{doctor visits} and \textit{FBwtested} may exhibit moderate performance, detecting 30-50\% of trends at the same window size, with higher delay. These signals tend to be more proximal to disease activity but with potential biases, such as the rise in telemedicine suppressing in-person visits \cite{patel2021variation} or unreported at-home testing limiting testing rate interpretability \cite{park2023unreported}. Weaker signals like \textit{QUIDELpositive} and \textit{FBwcli} typically recapitulate less than 20\% of trends, with greater than a month's delay. \textit{QUIDELpositive} measures the proportion of positive tests among those tested, which may be influenced by testing strategies, test availability, and gradual changes in healthcare-seeking behavior, decoupling them from the true incidence of infection \cite{chiu2021using}, \textit{FBwcli} captures self-reported symptoms, easily confounded by co-circulation of respiratory virsues \cite{chow2023effects}. For further analysis, we split the streams into 3 categories based on these results, the 3 ground truth streams with strong performance (GT), 5 streams with moderate performance (Medium), and 4 streams with weaker performance (Weak).

We perform 2 additional analyses to support our conclusions. First, we define ground truth using only CPR admissions, and observe similar patterns among signal strength, particularly that JHUcases and CHNG healthcare claims are still powerful data sources (Supplementary Fig. \ref{fig:one_gt}). Second, we assess alternate versions of LRTrend (Supplementary Fig. \ref{fig:alternate}). We use LRTrend's ground truth approach, but in real time, observing that the interpretable local regression is similarly powerful to the optimal parameter setting, without having instability due to the additional penalty parameter to tune. We additionally apply a commonly used moving average smoother instead of local regression to identify trend alarms, observing greatly reduced performance, indicating that local regression implicitly smooths trends better than a simple moving average. 

We conclude that LRTrend defines a consistent ground truth, powerfully detects sharp trend increases in near real-time, and that many disease-relevant data streams may be useful in this detection.


\subsection*{Aggregating across data streams for within-region denoising}

We explore aggregating the results of different data streams within a region with LRTrend's multivariate functionality, combining the $p$-values produced by each univariate stream (see Materials and Methods). Evaluations are performed using the same ground truth as previous, across 152 HRRs in the United States that has ground truth defined. We compare against Stolerman et al \cite{stolerman2023using}, another multivariate method which uses multiple data streams to identify regions of sharp increase, and the best performing univariate versions of LRTrend.

\begin{figure} 
	\centering
	\includegraphics[width=1.0\textwidth]{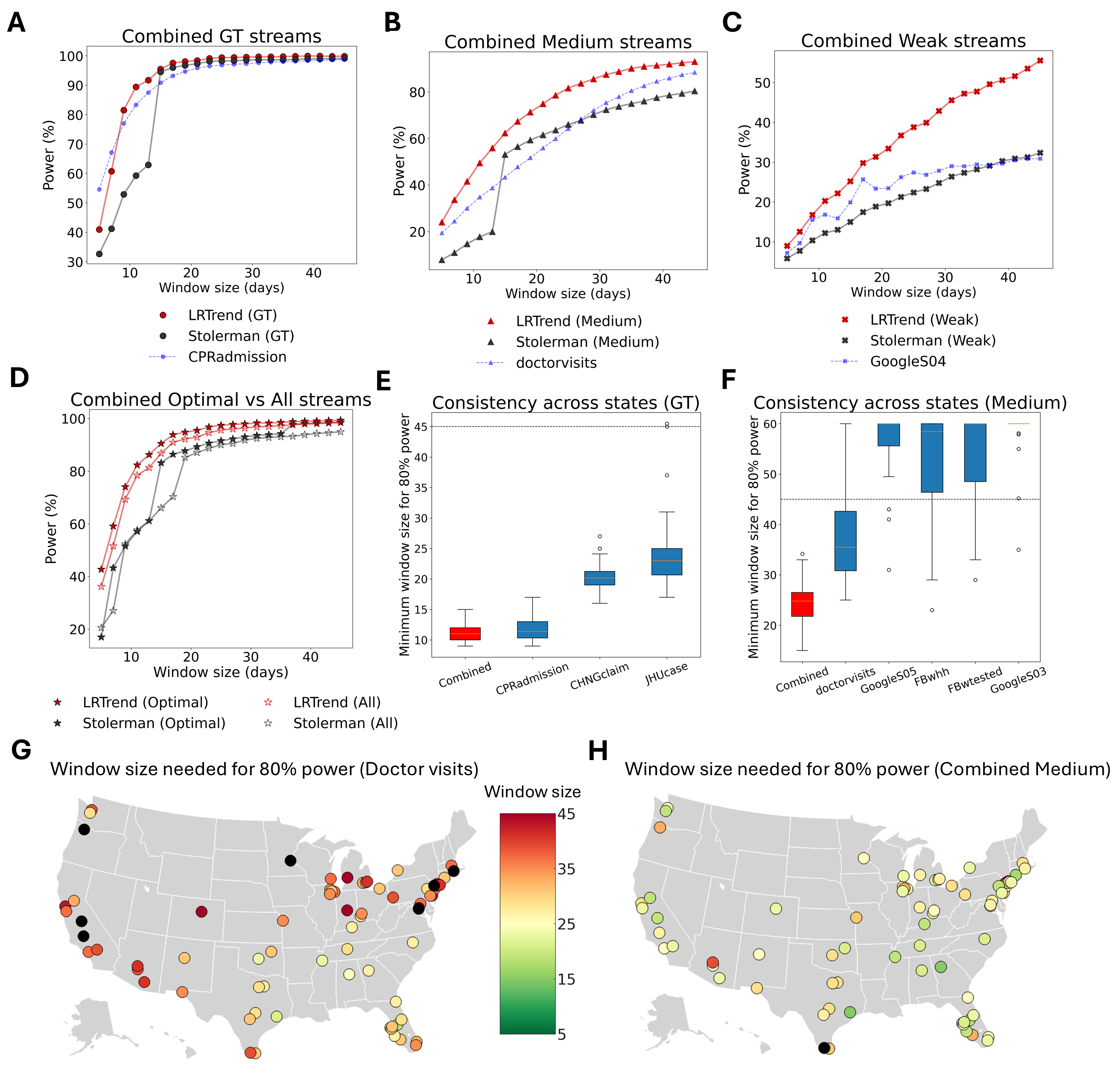}
\caption{\textbf{Multi-stream aggregation.} (\textbf{A–C}) Power for LRTrend and Stolerman using 3 GT, 5 Medium, and 4 Weak stream sets, respectively, compared to using LRTrend on each group's strongest individual stream. (\textbf{D}) Power for LRTrend and Stolerman’s optimal stream combination versus using all streams. (\textbf{E–F}) Average window size per state for 80\% power when combining GT and Medium streams versus individual streams (dashed line = maximum window size). (\textbf{G–H}) Window size needed for 80\% power across HRRs, mapped for doctor visits (G) and combined Medium streams (H).}
	\label{fig:multivariate}
\end{figure}

We aggregate all the GT, Medium, and Weak signals together and assess LRTrend's signal combination approach in Figs. \ref{fig:multivariate}A-C. We observe that for all 3 categories of signals, aggregating signals improves power at almost all window sizes compared to the top performing univariate stream of that category, (average power improvement 2\%, 21\%, 50\%, for GT, Medium, Weak, respectively). Consistent with the predominant idea that digital data streams may contain relevant information about disease activity \cite{salathe2012digital, santillana2016perspectives}, their proper combination can yield a more powerful test of trend significance. Notably, the improvement is greatest for the lower qualities of streams, indicating that in the absence of a single high quality measure of disease, many lower quality streams may be a good proxy. Compared to the method of Stolerman et al, LRTrend improves detection power, often substantially, using the same set of data streams (average power improvement 7\%, 30\%, 70\% for GT, Medium, Weak, respectively). Using the top performing stream with LRTrend may even surpass that of Stolerman et al's combined approach, particularly at window sizes less than 2 weeks where LRTrend's power can be 2 or 3-fold higher, likely due to the enhanced granularity from our smooth trend estimates. Similar results are observed for detection delay (Supplementary Fig. \ref{fig:combined_delay}).

We next analyze what constitutes a good combination of signals. We search over all possible combinations of signals and select the one with the top performance as our optimal, effectively overfitting the stream combination to this dataset. The resulting combination includes 10 data streams, only omitting the Weak category streams of \textit{FBwcli} and \textit{GoogleS04}. We similarly obtain Stolerman et al's optimal combination (which happens to be the same 10 streams), and also consider naively combining all signals for both methods. Results are reported in Fig. \ref{fig:multivariate}D. While there does exist an optimal combination of signals that differs from the set of all signals, the performance difference is negligible, with at most a 5\% loss in power when using all signals. This indicates LRTrend is robust to the specific set of aggregated streams and does not need extensive tuning of data streams to include - one can simply input the set of all data streams available and achieve reliable results. 

 
 We hypothesize that different streams may have region-specific biases which can be alleviated through LRTrend's combination. To test this, we evaluate what window size is needed to achieve 80\% power (and thus what resolution can most trends be detected at) for individual streams versus their combination for GT and Medium categories, averaging this window size within states to assess regional bias. We exclude Weak as most streams had less than 80\% power at all window sizes. Results are reported in Fig. \ref{fig:multivariate}E, F. For GT streams, the combined model tends to both require less window size than any individual stream, and has systematically lower variance across states compared to individual streams (combined variance = 3.2 vs. 4.4, 6.5, and 49.4 for \textit{CPRadmission}, \textit{CHNGclaim}, and \textit{JHUcase}, respectively). This is effect is even more pronounced for medium streams, likely due to the larger noise and lower initial power (combined variance = 17.7 versus 105.4, 51.4, 105.5, 90.7, and 25.9 for \textit{doctor visits}, \textit{GoogleS05}, \textit{FBwhh}, \textit{FBwtested}, and \textit{GoogleS03}, respectively). Visualizing this effect, we plot this performance for \textit{doctor visits} and the combined medium model in Fig. \ref{fig:multivariate}G, H. \textit{Doctor visits} alone does indeed have strong regional bias and performance is particularly low for much of the West Coast, Northeast, and Midwest regions. This may reflect regional discrepancies such as telemedicine practices suppressing in-person visits \cite{patel2021variation} or region-specific decoupling of infection from care seeking \cite{lewnard2022clinical}. However, after integration, most of these regions see a substantial reduction in window size needed, often greater than 2 weeks, and trend detection capability is now much more consistent across regions. While individual streams may have region-specific biases and noise, LRTrend's combination can counter the noise any individual stream may have and make the model consistent across all regions.


We conclude LRTrend can effectively and robustly integrate multiple low-powered data streams to create a denoised and powerful test of trend significance.

\subsection*{Learning useful epidemic networks for cross-region aggregation}

For a given data stream, we may have access to data from multiple geographic regions, in this case HRRs. We explore to what extent we can quantify the similarity between regions, and how we can aggregate information across regions to make a more powerful test. We construct epidemic neighbor graphs across all 305 HRR's by using the historical growth rates as features for time-series distance measurements (see Materials and Methods). We evaluate the power only for the 152 HRRs with ground truth defined. We compare this against aggregating across commonly used networks, including mobility information from the beginning of 2019 to the end of 2021 \cite{kang2020multiscale}, commuting flows from 2016-2020 5-Year ACS Commuting Flows \cite{ACS2020}, and simple geographic distance between regions. For each of our 12 data streams, we use LRTrend to learn a 3-NN epidemic neighbor graph between HRRs and average growth rates across this graph. We additionally assess limiting these neighbors to be in-state only to see the utility of within-state aggregation. We perform the same aggregation procedure with the mobility, commuting, and distance-based networks.

\begin{figure} 
	\centering
	\includegraphics[width=1.0\textwidth]{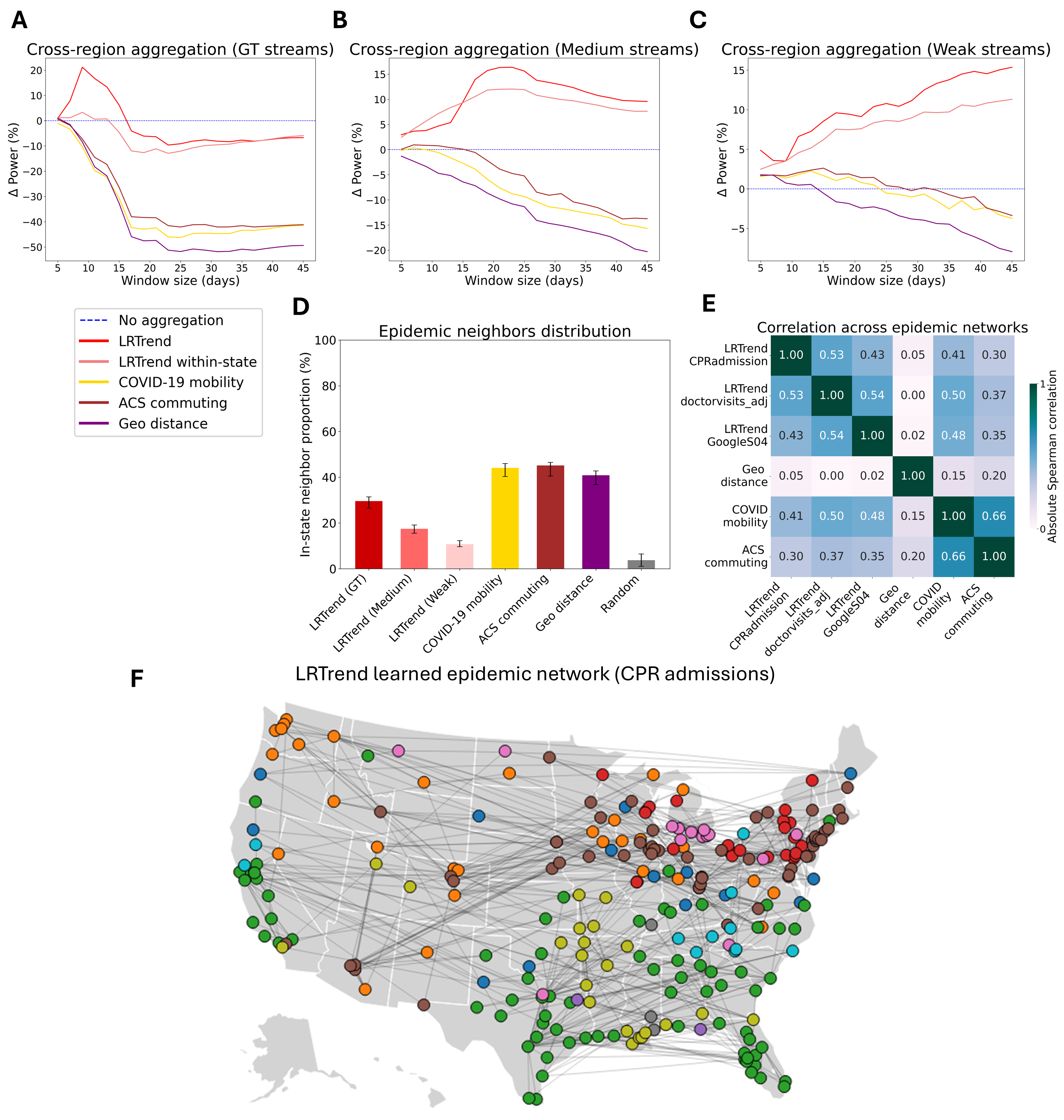} 

	\caption{\textbf{Geographic aggregation} (\textbf{A-C}) Change in power versus no aggregation averaged across streams with each aggregation method for GT, Medium, and Weak streams, respectively. (\textbf{D}) Average number of in-state neighbors, averaged across states with 95\% confidence intervals. (\textbf{E}) Absolute Spearman correlation in learned epidemic distances between each network. (\textbf{F}) 3-NN epidemic neighbor graph learned using LRTrend with CPR admissions. Regions are colored by cluster, lines represent 3-nearest neighbor connections.}
	\label{fig:geographic} 
\end{figure}

 Results averaged across GT, Medium, and Weak data streams are reported in Figs. \ref{fig:geographic}A, B, C. For all streams, LRTrend's aggregation significantly increases the power to detect significant trends, often by 10-20\%. For Medium and Weak streams, the improvement is more substantial, implying their lower power may be due to higher noise and improved more through cross-region denoising. We also note that despite the lower performance of some streams, such as those in the Weak category, they can still strongly identify similar neighbors to increase power. Interestingly, aggregating across mobility, distance, and commuting networks tend to actually reduce model performance, likely representing these networks alone are not good measures of epidemic similarity. Among these, distance is consistently the worst performing network. The optimal network representing similar epidemic dynamics is likely not defined simply by metrics proximal to human movement between location, but may be more complex multilayer networks that additionally encode policy and socio-demographic similarity, shared socioeconomic vulnerability, and common healthcare-resource utilization across HRRs \cite{balcan2009multiscale}. Supporting this, we observe that LRTrend aggregating across states often outperforms aggregating only within states, despite the shared policies, testing guidelines, and public-health reporting infrastructures that regions in the same state would be subject to. Results for delay is in Supplementary Fig. \ref{fig:geo_delay}.

We next analyze the qualities of LRTrend's learned networks. Observing the proportion of within-state neighbors for each trend (Fig. \ref{fig:geographic}D), around 10-30\% of 3-NN neighbors may be within the same state of each HRR, depending on signal quality. While higher than random (all one-sided $p<1^{-6}$), this is markedly lower than mobility, commuting, and distance based networks (all one-sided $p<1^{-6}$), which each have around 40\% within-state neighbors. Using the epidemic neighbor graphs inferred from the top performing streams of each category (\textit{CPR admissions}, \textit{doctor visits}, and \textit{GoogleS04}), we examine the Spearman's rank correlations between their learned distances in Fig. \ref{fig:geographic}E. LRTrend's learned distances are moderately correlated with each other across signal qualities (Spearman's r = 0.43-0.54), supporting that despite a given signal's lower quality, it can still construct a useful network. While LRTrend's distances are not very correlated with geographic distance (Spearman's r = 0.00-0.05), they are moderately correlated with mobility (Spearman's r = 0.41-0.50) and commuting flows (Spearman's r = 0.30-0.37). 
Human movement between locations likely captures some aspect of the LRTrend's learned epidemic network, consistent with previous work analyzing mobility \cite{badr2020association, chang2021mobility}, but other factors independent of mobility have a strong influence on what decides epidemic similarity, such as susceptibility differences between regions \cite{monod2021age}, differences in intervention effectiveness and adoption \cite{brauner2021inferring}, or climate and seasonal differences \cite{kissler2020projecting, baker2020susceptible}. Neighbor proportions for all data streams are available in Supplementary Fig. \ref{fig:in_state_all}.



We visualize the epidemic network LRTrend constructs using the CPR admissions data stream in Fig. \ref{fig:geographic}E, additionally applying K-means clustering (k=10) to identify groups of regions with epidemiological similarity. Expectedly, we observe HRR's surrounding a large city tend to belong to the same cluster, such as HRR's surrounding Chicago, Detroit, NYC, DC, and Seattle. 36.2\% of the 50 most populous cities' connections are within 150 miles. While most connections are reasonably close in distance, we observe it is almost never true that the 3 physically closest regions will be the epidemiological neighbors, consistent with our results as distance being the worst performing network for aggregation. Among the top 50 most populous cities, 17.4\% of their connections are to other major cities, often with strong regional and cultural connections, such as San Jose and Los Angeles, DC and Philadelphia, or Chicago and Minneapolis, consistent with observed spring waves largely affecting urban areas \cite{pei2021burden}. 
Among the 10 clusters, several notable patterns emerge. First, we recapitulate the 3 clusters identified in a similar analysis at the county level \cite{stolerman2023using}. Specifically, the Pacific Northwest spanning Seattle to Portland clusters with several regions such as Salt Lake City, Boise, Reno, and locations in the Midwest, potentially representing Washington state's distinct variant \cite{bedford2020cryptic}; California, Texas, and Florida regions cluster together, consistent with the summer waves affecting the southern half of the US \cite{pei2021burden}; and the Northeast from DC to Boston clusters together, a highly connected mega-region with strong observed epidemic connections \cite{lemieux2021phylogenetic}. Second, we observe other regional clusters, including the South from New Orleans to St. Louis, potentially representing the unique virus transmission as a result of Mardi Gras\cite{ZELLER20214939}. Third, by far the most diverse clustering happens around the Midwest from Minnesota to Ohio. For instance, there are 4 or more separate clusters in each of Wisconsin, Illinois, and Indiana. In Wisconsin, the county containing college town and state capitol Madison was found to have distinct viral patterns from the county containing urban center Milwaukee despite their proximity\cite{moreno2020revealing}, a pattern replicated in our learned network as these two do not cluster together. The Midwest's mosaic of urban centers, college towns, and rural areas may contribute to its strong epidemic heterogeneity \cite{ives2021estimating, unwin2020state}, and interdependence between these regions may further complicate epidemic response and spread \cite{pnashetero}. Similar epidemic networks for other data streams are available in Supplementary Fig. \ref{fig:geo_alt}.

In summary, LRTrend accurately identifies epidemic neighbors across the USA, and aggregating information across these neighbors can markedly denoise signal and increase model performance.

\section*{Discussion}

We have presented LRTrend, an interpretable machine learning framework to identify outbreaks in real time, given any health-related time series. LRTrend effectively incorporates information to increase detection power both within-region, through multiple signal aggregation, and across regions, through epidemic network learning and aggregation. Applying LRTrend at the HRR-level, we found it substantially outperforms related work in trend detection and learns more relevant epidemic networks. LRTrend serves as a simple and readily available method for detecting outbreaks, as well as a tool for quantifying epidemic similarity and constructing epidemic networks which may be useful for future pandemic response.

Our study has a few limitations. First, using alternate, less disease-specific data streams has the risk of detecting uptrends from other diseases, such as those infected with flu having similar Google search trends to those infected with COVID-19. The diverse data sources need to be present across multiple regions, and data collected in a comparable manner across states. We assume a static epidemic network structure, but region-specific policy responses could always alter the network dynamically. While we observe the historical FPR threshold tends to be comparable across time, a situation could always arise such that this is no longer true, such as drastic changes in data collection procedures.

Several future works are motivated by our results. First, additional, more complex models of epidemiological data could be incorporated, such as overdispersed Poisson or negative binomial. Second, epidemic networks could be dynamically modeled and updated, such as incorporating only the previous few months/years information, which may be able to capture sudden regional policy shifts. Third, a method for constructing the optimal aggregation strategy, searching over data streams and network parameters, could prove useful in maximizing detection power. The data considered here is only a subset of all potentially relevant streams, which will likely continue to grow in scale and diversity as more data is collected. Identifying the most relevant streams across hundreds of candidates could enable an extremely powerful and rapid outbreak detector.



\section*{Acknowledgments}
This material is based upon work supported by the United States of America Department of Health and Human Services, Centers for Disease Control and Prevention, under award numbers U01IP001121 and NU38FT000005; and contract number 75D30123C1590. Any opinions, findings, and conclusions or recommendations expressed in this material are those of the author(s) and do not necessarily reflect the views of the United States of America Department of Health and Human Services, Centers for Disease Control and Prevention. We thank Roni Rosenfeld for useful advice throughout the project and Martin Jinye Zhang for advice on manuscript preparation.

\paragraph*{Funding:}

R.L and B.W. were supported by the Delphi Influenza Forecasting Center of Excellence grant U01IP001121, Digital Public Health Surveillance for the 21st Century contract 75D30123C1590, and the The Delphi Center for Outbreak Analytics and Disease Modeling in Public Health Response: Innovation and Coordination grant NU38FT000005. A.T. was partially supported by a fellowship from Carnegie Mellon University’s Center for Machine Learning and Health.

\paragraph*{Author contributions:}

R.L and B.W. conceived the idea and constructed the methodology. R.L. and A.T. performed the analyses. R.L, B.W., and A.T. wrote the manuscript.

\paragraph*{Competing interests:}
There are no competing interests to declare.

\paragraph*{Data and materials availability:}

The LRTrend method and analysis code is publicly available at \href{https://github.com/Rachel-Lyu/LRTrend}{\texttt{https://github.com/Rachel-Lyu/LRTrend}}. Raw time-series data is available through the Delphi Epidata API \cite{reinhart2021open}.

\section*{Materials and Methods}


In the following sections we present a detailed explanation of our methods and data. In this work, we aim to estimate and detect changes in trends that reflect shifts in population health. Specifically, we aim to detect these trends in a real-time setting, assuming future information is not available at the time of analysis. This task is distinct from outlier detection, detecting short anomalies that revert to normal afterwards, and change point detection, which are abrupt, non-reverting changes. 

\subsection*{Data Sources}

Our data were accessed via the Delphi Epidata API \cite{reinhart2021open}, except Change Healthcare data which was extracted separately. The Facebook data is gathered by Delphi in partnership with Facebook. This survey collects responses from tens of thousands of Facebook users every day, asking them a broad set of COVID-related questions, including whether they or anyone in their household are currently experiencing COVID-related symptoms. The Google data contains the estimated volume of web searches related to COVID-19 and H5N1 highly-pathogenic avian influenza, showing the average relative frequency of each search in each location. 

\subsubsection*{COVID-19 Admissions}
\emph{CPR Admissions}: The COVID-19 Community Profile Report (CPR) is based on the daily report published by the Data Strategy and Execution Workgroup under the White House COVID-19 Team. It contains detailed daily-resolution figures on cases, deaths, testing, hospital admissions, healthcare resources, and vaccinations. We use all confirmed COVID-19 hospital admissions occurring each day, smoothed in time with a 7-day average.

\subsubsection*{JHU COVID-19 Cases}
\emph{JHU Cases}: Johns Hopkins University's Center for Systems Science and Engineering (JHU-CSSE) reports confirmed COVID-19 cases. This dataset provides daily counts of laboratory-confirmed COVID-19 cases.

\subsubsection*{Change Healthcare COVID-19 Claims}
\emph{CHNG Claims}: Change Healthcare is a healthcare technology company that aggregates medical claims data from many healthcare providers. This dataset includes aggregated counts of outpatient doctor visits with confirmed COVID-19, based on claims data that has been de-identified in accordance with HIPAA privacy regulations.

\subsubsection*{Doctor Visits}
\emph{Doctor Visits}: Information about outpatient visits provided by health system partners. Using outpatient claims counts, Delphi estimates the percentage of doctor's visits in a given location, on a given day, that are primarily about COVID-related symptoms, smoothed in time using a Gaussian linear smoother.

\subsubsection*{Google S03}
 \emph{GoogleS03}: Relative search frequency related to Fever, Hyperthermia, Chills, Shivering, and Low grade fever.
 
\subsubsection*{Google S04}
 \emph{GoogleS04}: Relative search frequency related to Shortness of breath, Wheeze, Croup, Pneumonia, Asthma, Crackles, Acute bronchitis, and Bronchitis

\subsubsection*{Google S05}
 \emph{GoogleS05}: Relative search frequency related to Anosmia, Dysgeusia, and Ageusia
 
\subsubsection*{QUIDEL positive}
\emph{Quidel Positive}: Percentage of antigen tests that were positive for COVID-19 (all ages), with no smoothing applied. This dataset is based on COVID-19 antigen tests provided by Quidel, Inc. An antigen test can detect parts of the virus that are present during an active infection. 

\subsubsection*{Facebook whh}
\emph{FB whh}: Estimated percentage of people reporting illness in their local community. Participants are asked if they know anyone in their local community who has COVID-like symptoms, defined as fever along with either cough, shortness of breath, or difficulty breathing.

\subsubsection*{Facebook wcli}
\emph{FB wcli}: Estimated percentage of people with COVID-like illness. Delphi estimates the percentage of people self-reporting COVID-like symptoms, defined as fever along with either cough, shortness of breath, or difficulty breathing.

\subsubsection*{Facebook wtested}
\emph{FB wtested}: Estimated percentage of people who were tested for COVID-19 in the past 14 days, regardless of their test result.

\subsubsection*{Facebook positive}
\emph{FB positive}: Estimated test positivity rate among people tested for COVID-19 in the past 14 days.

\subsection*{Local regression: identifying increasing trends in a short window}
To identify trends, we aim to identify a significant growth rate of an epidemic incidence $r(t)$, defined as $r(t) = \frac{y'(t)}{y(t)}$, where $y$ is a measure of epidemic incidence (e.g., counts, rates, log counts, etc.) and $y'(t)$ is the derivative of $y$ at time $t$. Taking the natural log, we can rewrite $r(t) = \frac{d}{dt}\ln(y(t))$. We approximate the derivative of $\ln(y(t))$ at time $t$ as $\ln(y_{t+1}) - \ln(y_t)$. This approach assumes a locally constant exponential increase or decrease, aligning with epidemiological understanding \cite{parag2022epidemic}, and correcting for asymmetry between past and future growth rates. 
While public health officials are often interested in the time-varying reproductive number $R(t)$, this metric is highly model-dependent, requiring assumptions about incidence and generation intervals. For example, in the linearized SIR model, $R(t) = 1 + \frac{r(t)}{\gamma}$, where $\gamma$ is the assumed recovery rate \cite{fraser2007estimating}. In the SEIR model, 2 parameters are assumed: $R(t) = \left(1 + \frac{r(t)}{\gamma_1}\right)\left(1 + \frac{r(t)}{\gamma_2}\right)$ \cite{wallinga2007generation}. In contrast, the growth rate $r(t)$ is directly observable and comparable across trends, enabling scale-invariant detection of trend significance.

Consider a time window of size $n$ with consecutive observations $y = \{ y_1, y_2, \ldots, y_n \}$ and time index $x = \{ 1, 2, \ldots, n \}$. Our goal is to provide an estimate of $r(t)$ given a window size, denoted as $\beta$, the average first-order difference in log space over this window. Traditional smoothing methods, such as splines and moving averages, often perform poorly at the ends of time series  \cite{jahja2022real}, making them unsuitable for real-time trend detection. We instead opt for local regression, and consider 3 separate models for estimating $\beta$, the proper choice of which depends on the observed distribution of the data.

\subsubsection*{Local linear regression}
The first model is linear regression, assuming $\log(y_t)$ is Gaussian distributed given $x_t$, i.e., $\log(y_t) \mid x_t \sim N(\alpha + x_t \beta, \sigma^2)$. This model assumes homogeneous multiplicative noise, such that in the original space, the noise magnitude is proportional to the counts, aligning with our observations of COVID-19 cases. The model becomes $\log(y_t) = \alpha + x_t \beta + \epsilon_t, \epsilon_t \sim N(0, \sigma^2)$. We further prove $\hat\beta \sim N(\beta, \frac{12\sigma^2}{n^3-n})$ (Supplementary Text), meaning the variance of estimator $\hat\beta$ shrinks at rate $\mathcal{O}(n^{-3})$ as $n$ increases, due to the fact that solving a linear equation with day indices as the regressor leads to a sum over day indices squared, a cubic function, and performance will rapidly increase with larger window size.

\subsubsection*{Local Poisson regression}

The second model is Poisson regression, assuming $y_t$ is Poisson distributed given $x_t$, i.e., $y_t \mid x_t \sim \text{Poi}(\mu_t)$ with $\log(\mu_t) = \alpha + x_t \beta$. This assumption matches influenza data well, where overdispersion is not an issue.

\subsubsection*{Local Negative Binomial regression}

The third model is Negative Binomial regression, assuming $y_t$ follows a Negative Binomial distribution given $x_t$, that is, $y_t \mid x_t \sim \text{NegBin}(\mu_t, r)$ with $\log(\mu_t) = \alpha + x_t \beta$. This model contains overdispersion parameter $r$, providing flexibility for real-world data where the variance exceeds the mean, such as COVID-19 data. To reduce complexity, we employ a plug-in estimator $\widehat{r}^{-1} = \frac{\hat{\mathbb{V}}(y_t) - \hat{\mathbb{E}}(\mu(x_t)) + \hat{\mathbb{E}}^2(\mu(x_t))}{\hat{\mathbb{E}}(\mu^2(x_t))} - 1$.


\subsection*{Constructing retrospective ground truths}

We develop a retrospective smoother that builds upon the \href{https://cmu-delphi.github.io/delphi-epidata/api/covidcast-signals/chng.html#day-of-week-adjustment}{Delphi Epidata API}  and Rumack et al \cite{rumack2023modeling}. The original smoother is a Poisson model, but we observe several count-based data streams exhibit overdispersion. To address this, we expand the Poisson framework to include Log-Normal smoothing, as well as a multivariate extension to accommodate multiple time series. This smoother indicates ground truth increasing trends, which real time methods can then aim to recapitulate using only real time information.
\subsubsection*{Multi-stream retrospective smoother}
Here we assume Poisson or Log-Normal distributed counts $y_t$ with mean parameter $\mu_t$, and that the weekday effect is multiplicative. Formally, 
\begin{equation}
\log \mu_t = \alpha_{\text{wd}(t)} + \log \phi_t
\end{equation}
where $\text{wd}(t) \in \{0, \dots, 6\}$ is the day-of-week of time $t$, $\alpha_{\text{wd}(t)}$ is the corresponding weekday correction, and $\phi_t$ is the latent disease burden at time $t$. To fit the $\alpha$ parameters, we minimize the following convex objective function: 
\begin{equation}
f(\alpha, \phi | y) = -\log \ell (\alpha,\phi|y) + \lambda ||\Delta^3 \phi||_1
\end{equation}
$\Delta^3 \phi$ is the third difference of $\phi$. For identifiability, we constrain the sum of $\alpha$ to be zero by setting Sunday’s fixed effect to be the negative sum of the other weekdays. The penalty term $\lambda$ encourages the $\phi$ curve to be smooth and produce meaningful $\alpha$ values. More details of Poisson and Log-Normal smoothing are put in the Supplementary Text.

\begin{figure}
    \centering
    \includegraphics[width=0.9\linewidth]{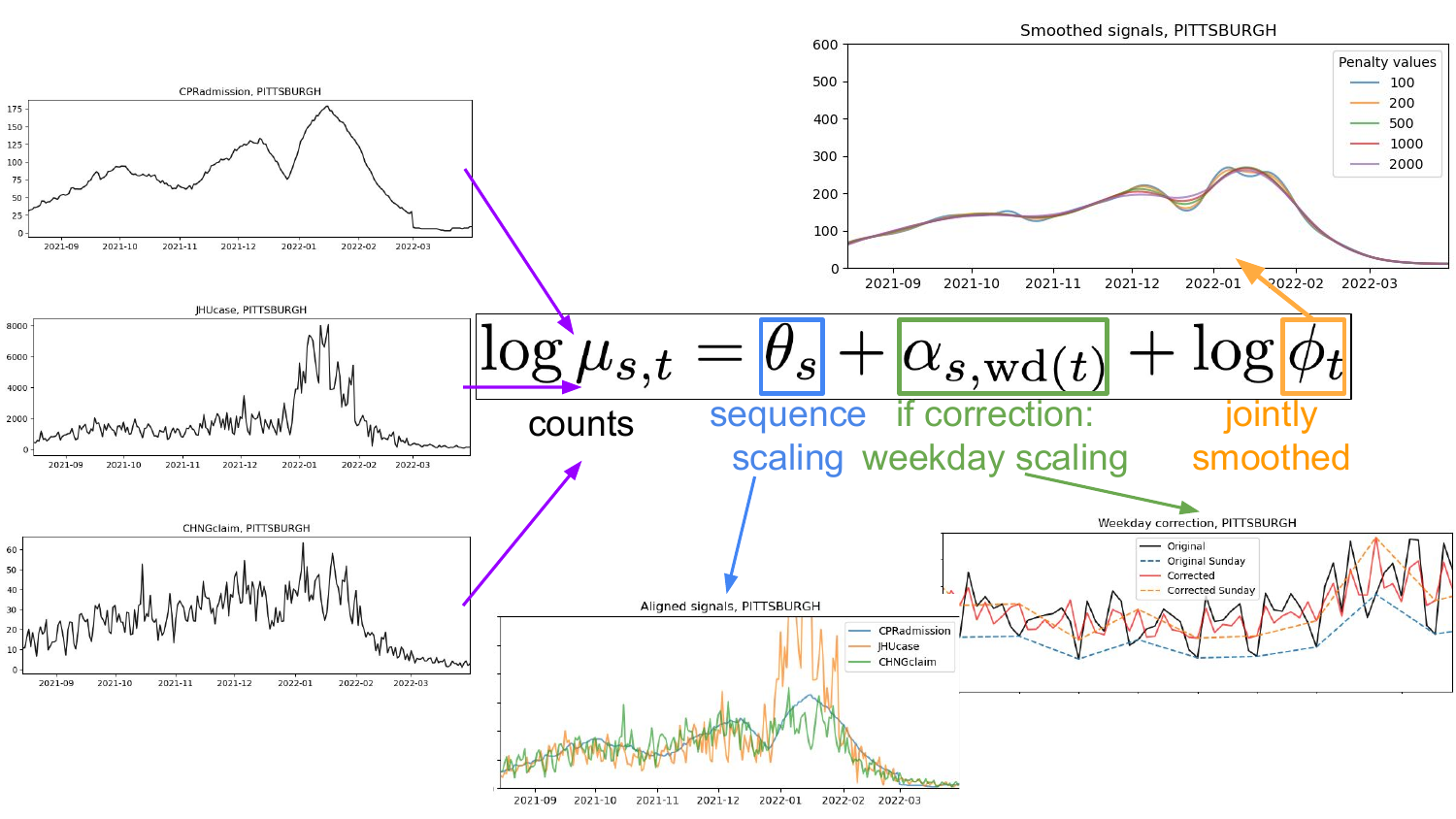}
    \caption{Flowchart of multivariate smoothing. Multiple input data sources are input, scaled, weekday corrected, and jointly smoothed to produce a single measure of latent disease burden.}
    \label{fig:flowchart}
\end{figure}

We extend our model to a multivariate setting to accommodate multiple time series, indexed by $s$. Each series is modeled using the following formulation:
\begin{equation}
\log \mu_{s, t} = \theta_s + \alpha_{s, \text{wd}(t)} + \log \phi_t
\end{equation}
where $\mu_{s, t}$ is the expected count for series $s$ at time $t$; $\theta_s$ is a series-specific scaling factor to align the magnitudes of different sequences; $\alpha_{s, \text{wd}(t)}$ represents the weekday effect for series $s$ on day $t$ (when we choose to correct that signal); $\phi_t$ is a latent common trend component shared across all series. We do not enforce that all series $y_s$ follow the same distributional assumptions; instead, we model each series separately according to its characteristics and $\mu_{s, t}$ can be computed from different models. If a series exhibits overdispersion, we may choose Log-Normal, whereas for another series with Poisson-like counts, we may opt for the Poisson model. By including $\log \phi_t$ as a common term, we allow the different series to share information about the underlying epidemic trend while still accounting for series-specific characteristics through $\theta_s$ and correction. The diagram of the overall process of the multivariate smoothing is shown as Fig. \ref{fig:flowchart}. 

\subsubsection*{Defining ground truth}
To create ground truth data for evaluation, we apply the smoother to the observed data using varied parameters ($\alpha$ values), generating multiple smoothed versions of each series with different penalties. We then estimate the growth rate for each smoothed series to identify consensus periods where a trend is increasing or not increasing, indicating robust evidence of a trend. All smoothed series must be in agreement about the direction of the trend.

To evaluate the performance of local regression, we first apply retrospective smoothing on multiple high-quality signals, annotating the null / alternative time intervals. We then locally regress all signals and obtain the growth rates. To compute accuracy at a desired False Positive Rate (FPR), we select the minimum growth rates estimated in local regression such that $FPR\%$ of historical trends indicated are false positives. This minimum growth rate is then used as a cutoff real-time trend detection. We measure the performance of the near real-time local regression coefficients to identify these periods. We evaluate performance with both power (correctly identifying increasing vs. non-increasing trends) and delay time between the actual onset of a trend and its detection. The default delay for an undetected trend is set to 60 days. 

\subsection*{Multiple signal combination}

Combining different signals in the same geographic region could reduce noise and increase statistical power. Local regression's coefficients may not be comparable across streams due to the magnitude difference between streams (e.g., rates between 0-1 and counts up to the thousands), however, the empirical quantiles and $p$-values are. Under the null hypothesis, the $p$-values for any stream are uniformly distributed between 0 and 1 (proof below). Thus, we propose that we can aggregate $p$-values across streams within the same region to increase statistical power.


\subsubsection*{Proof of uniformity under null hypothesis}

The local regression coefficients are asymptotically Gaussian distributed due to the Central Limit Theorem and the properties of maximum likelihood estimators. Specifically, under regularity conditions, the estimators are consistent and asymptotically normal with a mean equal to the true parameter and variance equal to the inverse of the Fisher information matrix. Mathematically, for the estimated growth rate $\hat{\beta}$,  this asymptotic behavior is expressed as $\sqrt{n} (\hat{\beta} - \beta) \xrightarrow{d} N(0, I^{-1}(\beta))$, where $n$ denotes the sample size (i.e., the window size in our local regression), and $I(\beta)$ is the Fisher information evaluated at $\beta$. The derivations supporting this result for each regression method are provided in the Supplementary Text.

\subsubsection*{Combining $p$-values within a region}

To effectively combine multiple signals, we adopt Stouffer's method \cite{stouffer1949american}, which is particularly suitable for aggregating $p$-values from independent tests when the test statistics approximate a normal distribution \cite{lyu2025federated}. Stouffer's method involves converting each $p$-value $p_i$ into a corresponding $z$-score $Z_i$ using the inverse cumulative distribution function $\Phi^{-1}$ of the standard normal distribution, $Z_i = \Phi^{-1}(1 - p_i)$. Stouffer's method then aggregates these $z$-scores into a single combined $z$-score $Z$ using a weighted sum $Z = \frac{\sum_{i=1}^k w_i Z_i}{\sqrt{\sum_{i=1}^k w_i^2}}$, where $k$ is the number of signals being combined and $w_i$ is a weight (set to $1$ under our equal weighting). Under the null hypothesis, where all individual $p$-values are uniformly distributed and independent, the combined $z$-score follows a standard normal distribution $Z \sim N(0, 1)$, so the combined $p$-value can be computed as $1 - \Phi(Z)$. The rationale behind this method is that by aggregating the 
$z$-scores, we effectively combine the evidence from multiple independent tests, reducing the variance and increasing the signal-to-noise ratio of the aggregated estimators compared to individual ones, enhancing the overall statistical power, and being robust to outliers.

\subsection*{Geographic smoothing}
While geographic proximity often correlates with epidemic trajectory, recent work \cite{thivierge2024does} demonstrated that the incorporation of strictly neighboring regions' information did not perform better than incorporating the national average. Geographical proximity doesn't always lead to similarity in disease transmission\cite{mcmahon2022spatial, mcmahon2023effect}. Instead, we directly compute epidemic similarity between regions using historical local regression coefficients as features, and aggregate across our learned epidemic network.

\subsubsection*{Learning an epidemic network}

While different geographic locations' counts/rates of a specified signal may be greatly varied, we observed their local regression coefficients remain comparable, and this is consistent across both different regions and time points. This enables the use of distance metrics to identify the nearest neighbors in epidemic trajectory using LRTrend's regression coefficients.

To compute the distance, we employ soft Dynamic Time Warping (soft-DTW) \cite{cuturi2017soft}, a differentiable variant of the traditional DTW algorithm. The soft-DTW distance between two time series (coefficient series here) $\beta_{1} = \{\beta_{1, 1}, \beta_{1, 2}, \ldots, \beta_{1, T_1} \}$ and $\beta_{2} = \{ \beta_{2, 1}, \beta_{2, 2}, \ldots, \beta_{2, T_2} \}$ is defined as 
\begin{equation}
\text{soft-DTW}(\beta_{1, t}, \beta_{2, t}) = -\gamma \log \left( \sum_{P \in \mathcal{P}} \exp\left(-\frac{d(P, \beta_{1, t}, \beta_{2, t})}{\gamma}\right) \right),
\end{equation}
where $\mathcal{P}$ is the set of all possible warping paths, $d(P, \beta_{1, t}, \beta_{1, t})$ is the cumulative distance along path $P$, and $\gamma > 0$ is a smoothing parameter to control the softness of alignment. 

\subsubsection*{Aggregating with an epidemic network}

We identify a set of $k$ nearest neighbors $\mathcal{N}_s$ for each target series $s$ using our learned distances. We opt for setting $k=3$, as the inclusion of more neighbors had minor impact. We aggregate the local regression coefficients $\{ \hat\beta_s \}_{s \in \mathcal{N}_s}$ as a weighted average of the coefficients from its neighbors $\hat\beta_{s, agg} = \frac{\sum_{s \in \mathcal{N}_s} w_s \hat\beta_s}{\sum_{s \in \mathcal{N}_s} w_s}$, where $w_s$ is the weight. We simply set all weights to $1$ here. 

\bibliographystyle{sciencemag}
\bibliography{science_template}

\newpage


\renewcommand{\thefigure}{S\arabic{figure}}
\renewcommand{\thetable}{S\arabic{table}}
\renewcommand{\theequation}{S\arabic{equation}}
\renewcommand{\thepage}{S\arabic{page}}
\setcounter{figure}{0}
\setcounter{table}{0}
\setcounter{equation}{0}
\setcounter{page}{1} 


\begin{center}
\section*{Supplementary Materials for\\ \scititle}

Ruiqi~Lyu$^{1\ast}$,
Alistair~Turcan$^{1}$,
Bryan~Wilder$^{1}$\\
\small$^\ast$Corresponding author. Email: ruiqil@cs.cmu.edu

\end{center}


\subsubsection*{This PDF file includes:}
Figures S1 to S7\\
Supplementary Text

\clearpage
\begin{figure} 
	\centering
	\includegraphics[width=1.0\textwidth]{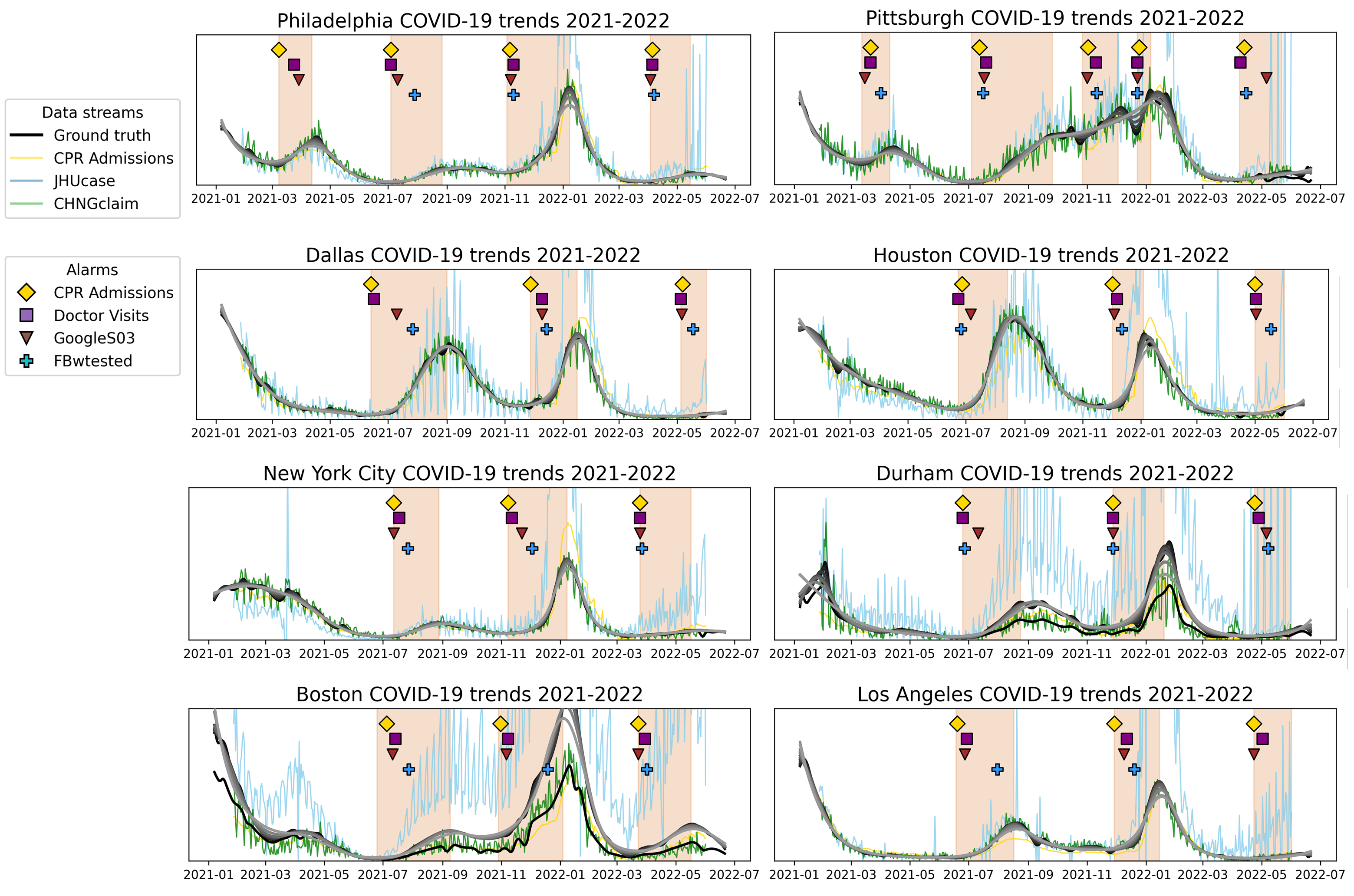}
	\caption{\textbf{Visualization for additional HRR's} Raw COVID-19 data streams CPR admissions, JHU cases, and Change Healthcare Claims. LRTrend's retrospective ground truth is indicated with a solid gray/black line, colored differently for different penalty values. Consensus outbreak regions are annotated in red shaded areas. Alarms are annotated from applying LRTrend with each data stream. }
	\label{fig:time_streams}
\end{figure}

\newpage
\begin{figure} 
	\centering
	\includegraphics[width=1.0\textwidth]{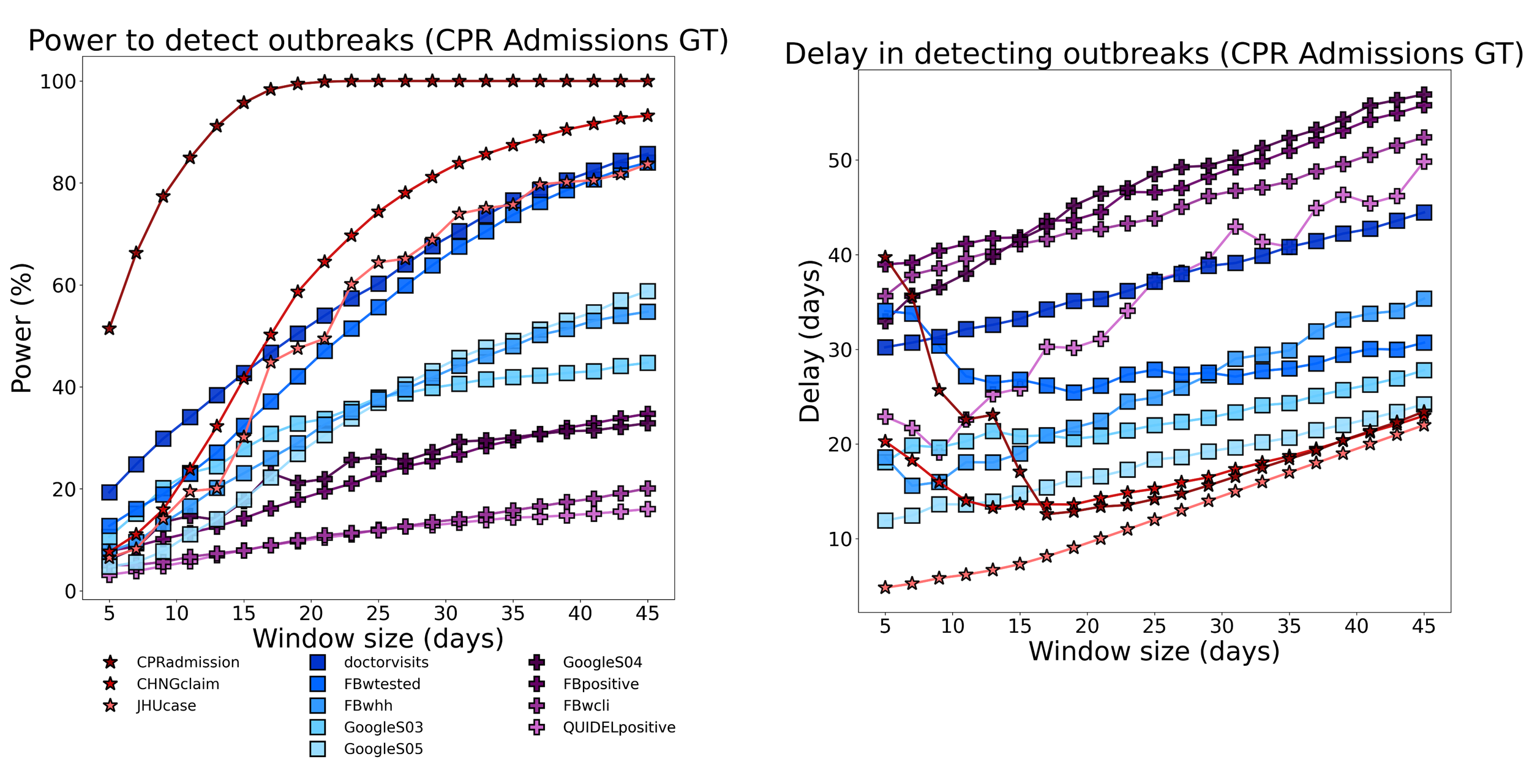}
	\caption{\textbf{Power and delay using CPR admissions only as ground truth} LRTrend's power and delay in detecting outbreaks in conjunction with each data stream versus window size used for detection. GT streams are colored red, Medium streams colored blue, and Weak streams colored purple.}
	\label{fig:one_gt}
\end{figure}
\clearpage
\begin{figure} 
	\centering
	\includegraphics[width=1.0\textwidth]{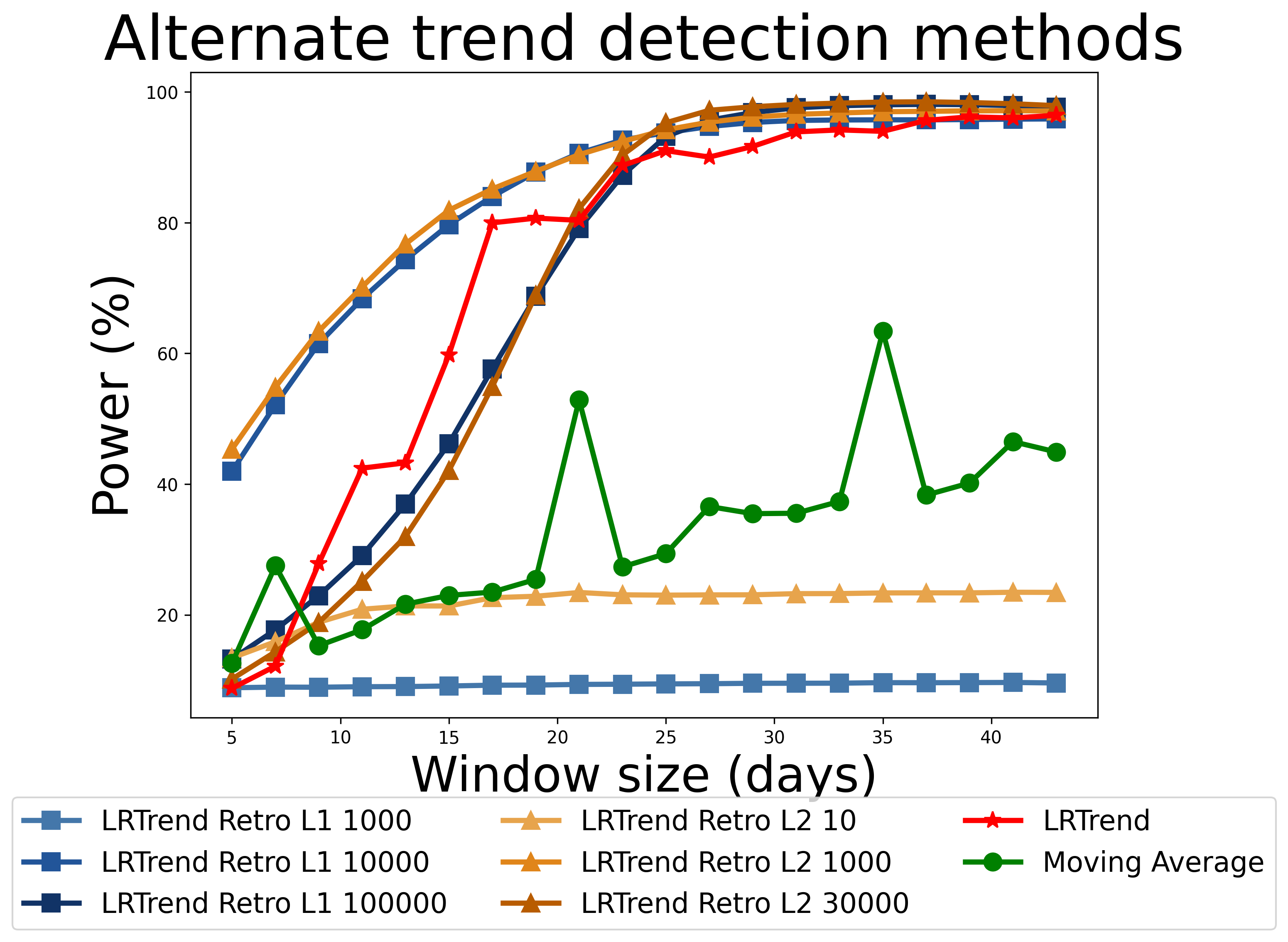}
	\caption{\textbf{Power using alternate versions of LRTrend on JHU cases} Power vs window size using JHUCases for LRTrend's retrospective method with L1 filtering (LRTrend Retro L1) and penalties 1000, 10000, and 10000, LRTrend's retrospective method with L2 filtering (LRTrend Retro L2) and penalties 10, 1000, 30000, default LRTrend, and a moving average smoother.}
	\label{fig:alternate}
\end{figure}
\clearpage
\begin{figure} 
	\centering
	\includegraphics[width=1.0\textwidth]{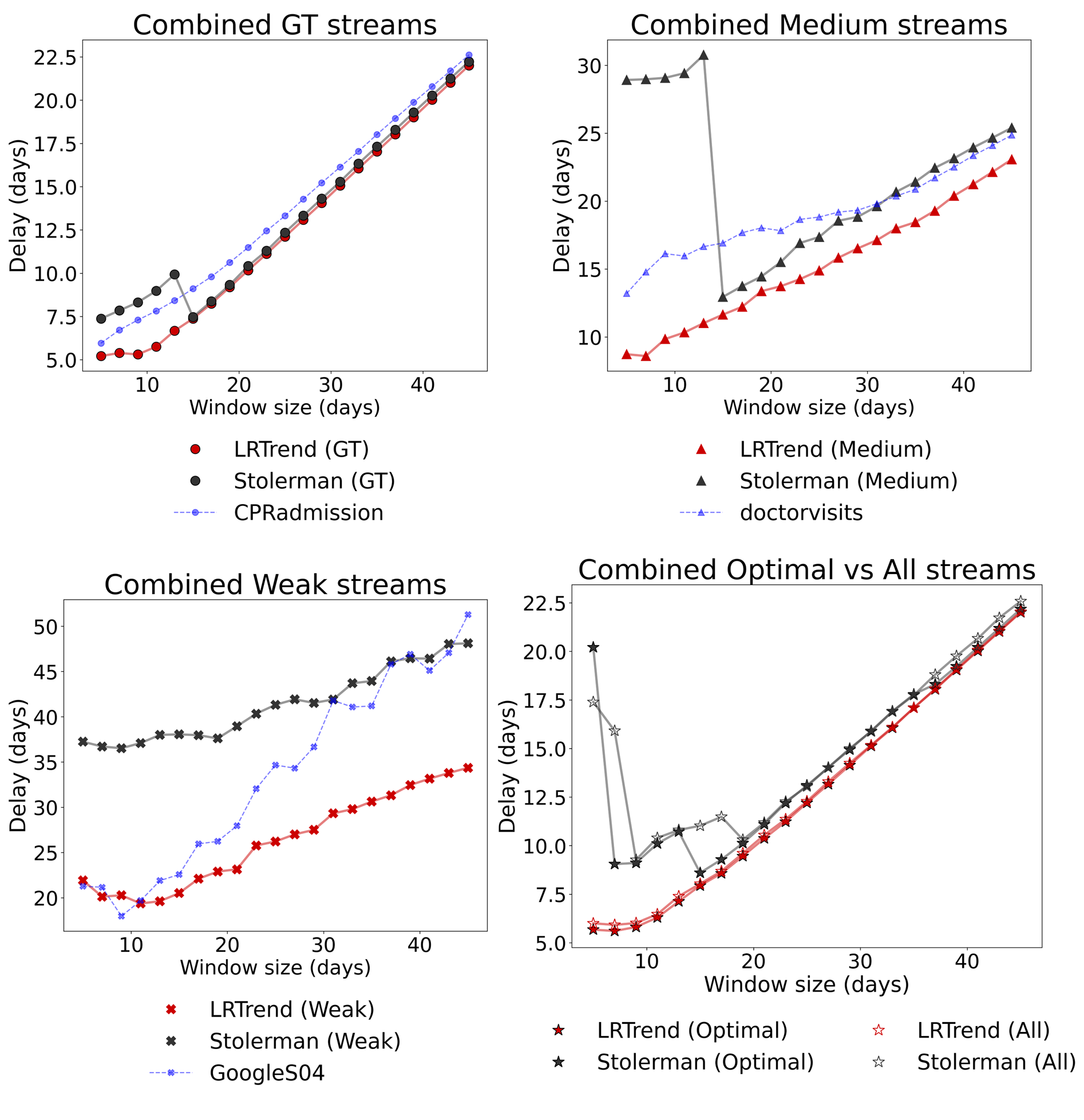}
	\caption{\textbf{Multi-stream aggregation delay} Delay for LRTrend and Stolerman using 3 GT, 5 Medium, 4 Weak, and optimal vs all stream sets, respectively, compared to each group's strongest individual stream. }
	\label{fig:combined_delay}
\end{figure}

\clearpage
\begin{figure} 
	\centering
	\includegraphics[width=1.0\textwidth]{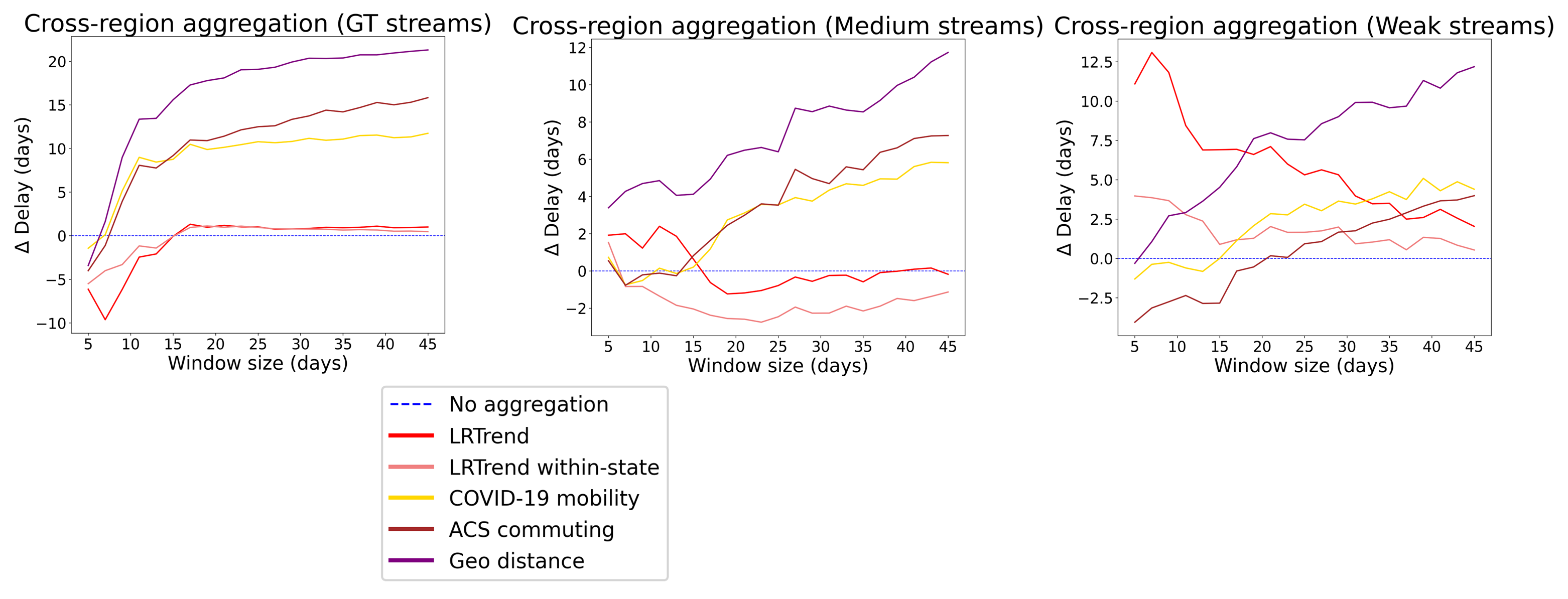}
	\caption{\textbf{Geographic aggregation delay}  Change in delay versus no aggregation averaged across streams with each aggregation method for GT, Medium, and Weak streams, respectively. }
	\label{fig:geo_delay}
\end{figure}


\clearpage
\begin{figure} 
	\centering
	\includegraphics[width=1.0\textwidth]{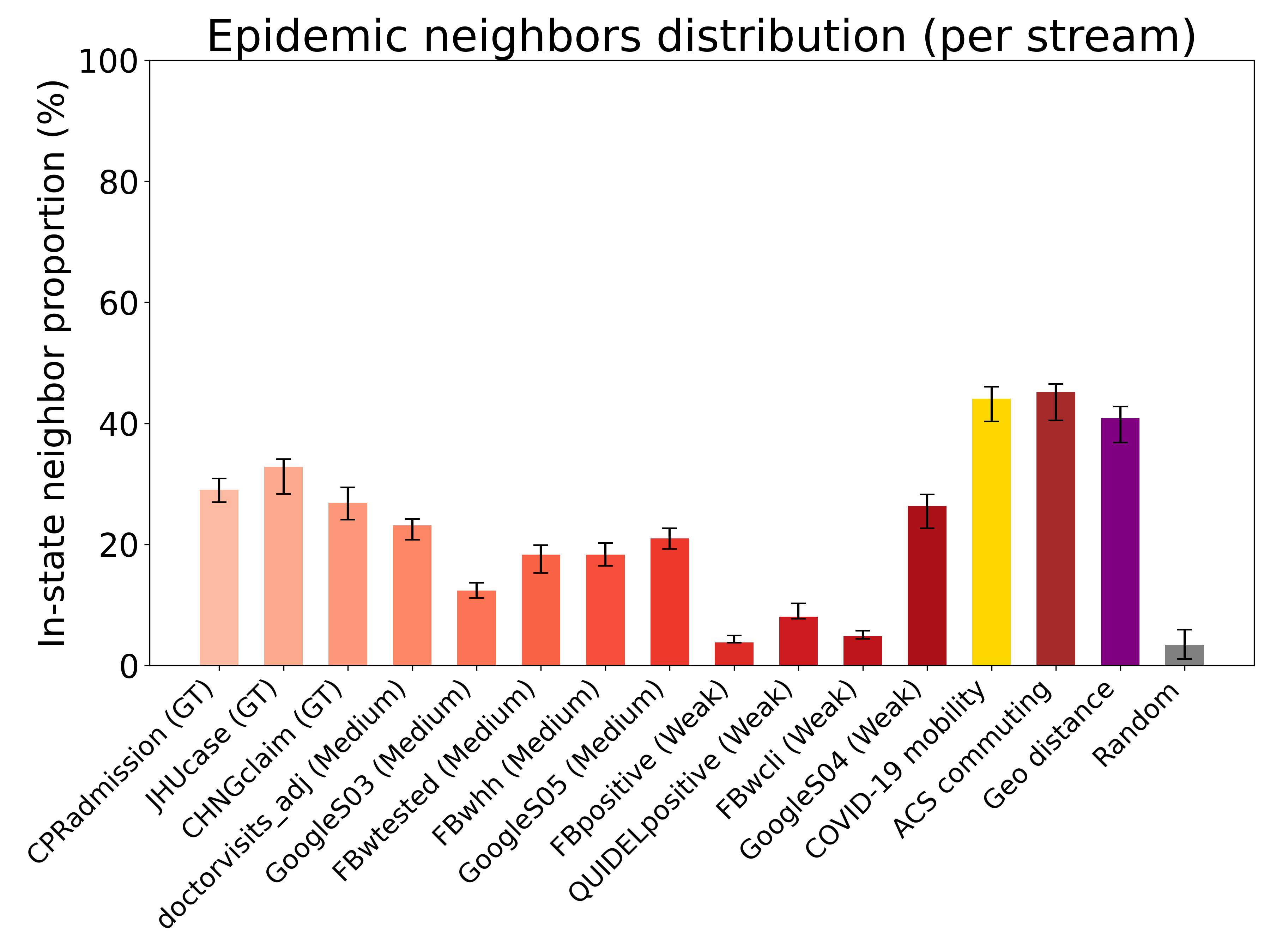}
	\caption{\textbf{In-state proportion of neighbors for individual streams} Average number of in-state neighbors, averaged across states with 95\% confidence intervals.}
	\label{fig:in_state_all}
\end{figure}

\clearpage
\begin{figure} 
	\centering
	\includegraphics[width=0.8\textwidth]{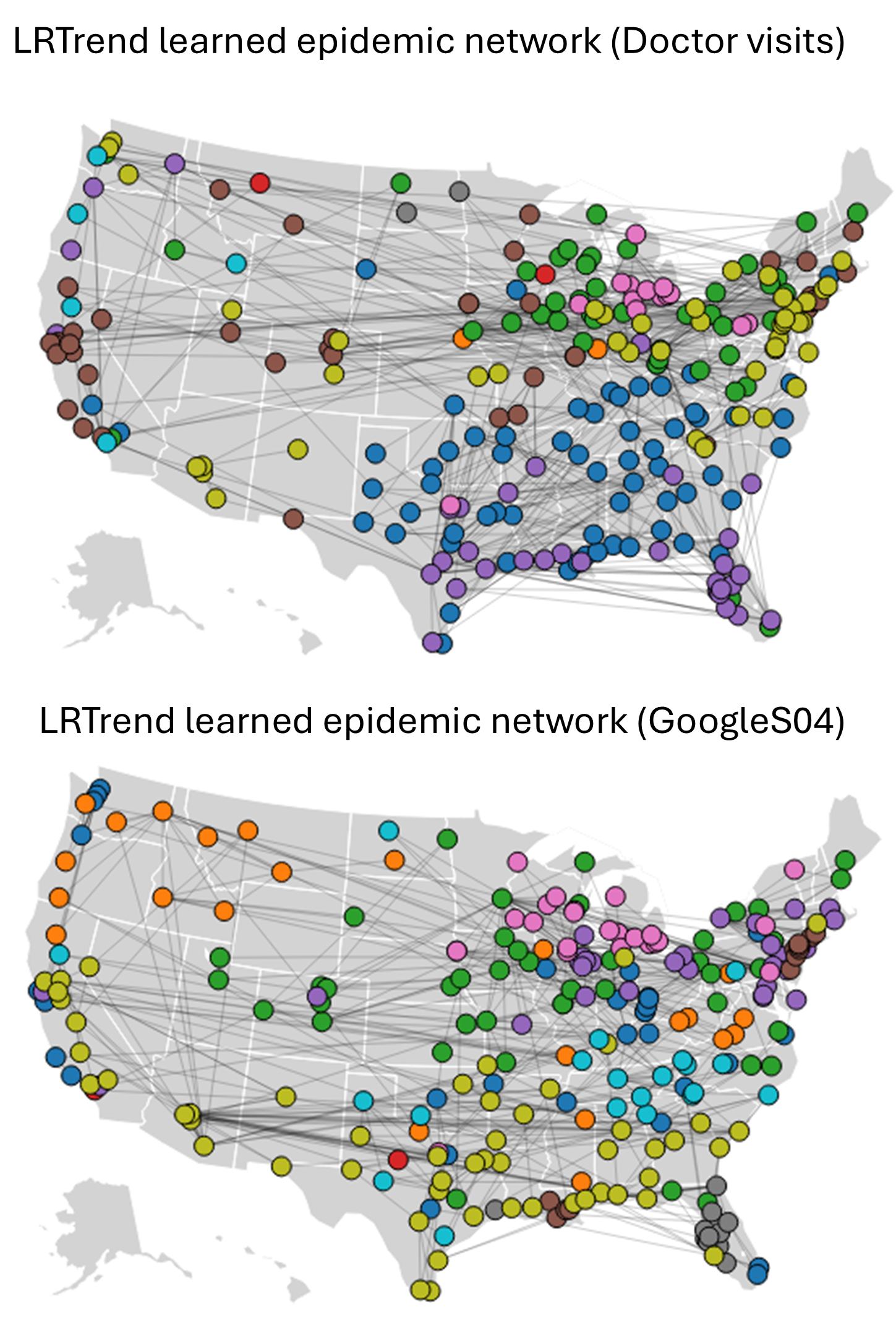}
	\caption{\textbf{Epidemic network for 2 additional data streams} 3-NN epidemic neighbor graph learned using LRTrend with Doctor visits and GoogleS04. Regions are colored by cluster, lines represent 3-nearest neighbor connections.}
	\label{fig:geo_alt}
\end{figure}

\clearpage

\subsection*{Supplementary Text}


\subsubsection*{Linear regression on log-transformed data}

The linear regression model is $\log(y_t) = \alpha + x_t\beta + \epsilon_t, \epsilon_t \sim N(0, \sigma^2)$. This naive linear regression has a close-form ordinary least squares solution. 

\begin{align*}
\begin{bmatrix}
\log(y_1) \\
\log(y_2) \\
\vdots \\
\log(y_n)
\end{bmatrix} &= \begin{bmatrix}
1 & 1 \\
1 & 2 \\
\vdots & \vdots \\
1 & n
\end{bmatrix} \begin{bmatrix}
\alpha\\
\beta
\end{bmatrix}\\
\end{align*}
Define
\begin{align*}
\overline X &= \begin{bmatrix}
1 & 1 \\
1 & 2 \\
\vdots & \vdots \\
1 & n
\end{bmatrix}, \overline{X}^{\top}\overline X = \begin{bmatrix}
n & \sum_{i=1}^n i \\
\sum_{i=1}^n i & \sum_{i=1}^n i^2
\end{bmatrix} = \begin{bmatrix}
n & \frac{n(n+1)}{2} \\
\frac{n(n+1)}{2} & \frac{n(n+1)(2n+1)}{6}
\end{bmatrix}\\
(\overline{X}^{\top}\overline X)^{-1} & = \frac{12}{n^2 (n+1)(n-1)} \begin{bmatrix}
\frac{n(n+1)(2n+1)}{6} & -\frac{n(n+1)}{2} \\
-\frac{n(n+1)}{2} & n
\end{bmatrix}
= \begin{bmatrix}
\frac{2(2n+1)}{n(n-1)} & -\frac{6}{n(n-1)} \\
-\frac{6}{n(n-1)} & \frac{12}{n(n+1)(n-1)}
\end{bmatrix}\\
\begin{bmatrix}
\hat\alpha\\
\hat\beta
\end{bmatrix} & = (\overline{X}^{\top}\overline X)^{-1} \overline{X}^{\top}\begin{bmatrix}
\log(y_1) \\
\log(y_2) \\
\vdots \\
\log(y_n)
\end{bmatrix} = \frac{12}{n^2 (n+1)(n-1)} \begin{bmatrix}
\frac{n(n+1)(2n+1)}{6} & -\frac{n(n+1)}{2} \\
-\frac{n(n+1)}{2} & n
\end{bmatrix}\begin{bmatrix}
\sum_{i=1}^n \log(y_i) \\
\sum_{i=1}^n i \log(y_i)
\end{bmatrix}\\
\mathbb{E}(\begin{bmatrix}
\hat\alpha\\
\hat\beta
\end{bmatrix}) &= \begin{bmatrix}
\alpha\\
\beta
\end{bmatrix}, 
\hat\beta = -\frac{6}{n(n-1)} \sum_{i=1}^n \log(y_i) + \frac{12}{n(n+1)(n-1)} \sum_{i=1}^n i \log(y_i)\\
\end{align*}
As $\log(y) | x \sim N(\alpha + x\beta, \sigma^2 I_n)$
\begin{align*}
\mathbb{E}&\left(\left(\begin{bmatrix}
\hat\alpha\\
\hat\beta
\end{bmatrix} - \begin{bmatrix}
\alpha\\
\beta
\end{bmatrix}\right)\left(\begin{bmatrix}
\hat\alpha\\
\hat\beta
\end{bmatrix} - \begin{bmatrix}
\alpha\\
\beta
\end{bmatrix}\right)^{\top}\right) = \sigma^2(\overline{X}^{\top}\overline X)^{-1}\\
\mathbb{V}(\hat\beta) &= \frac{12\sigma^2}{n(n+1)(n-1)}, \hat\beta \sim N(\beta, \frac{12\sigma^2}{n^3-n})
\end{align*}
We can see the variance of the unbiased growth rate estimator $\hat\beta$ shrinks with the increase of window size in the rate of $\mathcal{O}(n^{-3})$, which is vary rapid. From another perspective, the score function of this model is $U(\beta) = \frac{1}{\sigma^2} \sum_{i=1}^n (\log(y_i) - \alpha - i\beta) i$, meaning we are approximating $\log(y_i)$ with $\log(\mu_i)$, rather than approximating $y_i$ with $\mu_i$. Performing analysis in log space make this regression robust to extremely large counts outliers (which happens when the data revision is directly implemented on the accumulated counts on the day of findings). However, it will introduce additional error because of the logarithm and 0 counts will be a tricky situation. The Fisher information is $I(\begin{bmatrix}
\alpha\\
\beta
\end{bmatrix}) = \frac{1}{\sigma^2} \begin{bmatrix}
n & \frac{n(n+1)}{2} \\
\frac{n(n+1)}{2} & \frac{n(n+1)(2n+1)}{6}
\end{bmatrix}$. Variances can be inferred by the diagonal elements of the inverse of the Fisher information matrix so $\mathbb{V}(\hat\beta) = \frac{12\sigma^2}{n (n+1)(n-1)}$. 

\subsubsection*{Poisson regression}
The Poisson regression model assume the dependent variable $y_t$ is Poisson distributed given $x_t$: $y_t | x_t \sim \text{Poi}(\mu(x_t)$. The assumption is the counts are Poisson distributed where both expectation and variance equal to the same parameter $\mu$, meaning the noise magnitude is proportional to square root of the counts. Though this assumption doesn't fit into our observation of COVID cases well, as overdispersion was observed, if fits in flu data before COVID well. There's only one parameter to control expectation and variance, the simplicity of model is preferred in application. Also, the Poisson model is able to handle moderate amount of zero counts. The regression model is $y_t \sim \text{Poi}(\mu_t)$ where $\log(\mu_t) = \alpha + x_t\beta, \mu_t = e^{\alpha + x_t\beta}$. The score function $U(\beta) = \sum_{i=1}^n (y_i - e^{\alpha + i\beta}) i$. The Fisher information is $I(\begin{bmatrix}
\hat\alpha\\
\hat\beta
\end{bmatrix}) = \sum_{i=1}^n e^{\alpha + i\beta} \begin{bmatrix}
1 & i \\
i & i^2
\end{bmatrix}$.
When $\beta \neq 0$, 
\begin{align*}
    \det(I) &= \frac{e^{2\alpha + \beta}\left(-(n-1)e^{(2n+2)\beta} - n^2e^{(n+3)\beta} + (2n^2 - 2)e^{(n+2)\beta} - (n^2 - 1)e^{(n+1)\beta} - e^{n\beta} + e^{2\beta} - e^{\beta} + 1\right)}{(e^{\beta} - 1)^4}
\end{align*} 

When $\beta = 0$, $I(\begin{bmatrix}
\hat\alpha\\
\hat\beta
\end{bmatrix}) = \sum_{i=1}^n e^{\alpha} \begin{bmatrix}
1 & i \\
i & i^2
\end{bmatrix} = e^{\alpha} \begin{bmatrix}
n & \frac{n(n+1)}{2} \\
\frac{n(n+1)}{2} & \frac{n(n+1)(2n+1)}{6}
\end{bmatrix}$

The lower bound of the variances of the parameter estimates in Poisson regression can be inferred from the diagonal elements of the inverse of the Fisher information matrix, employing the Cramér-Rao Lower Bound (CRLB), which states that the covariance matrix of any unbiased estimator is at least as large as the inverse of the Fisher information matrix. When $\beta = 0$, $\mathbb{V}(\hat\beta) \geq \frac{12}{e^{\alpha} n (n+1)(n-1)}$; when $\beta \neq 0$, $$\mathbb{V}(\hat\beta) \geq \frac{(e^{n\beta} - 1)(e^{\beta} - 1)^3}{e^{\alpha }\left(-(n-1)e^{(2n+2)\beta} - n^2e^{(n+3)\beta} + (2n^2 - 2)e^{(n+2)\beta} - (n^2 - 1)e^{(n+1)\beta} - e^{n\beta} + e^{2\beta} - e^{\beta} + 1\right)}$$ The absolute value of $\beta$ is often small and the exact shrinking rate depends on its magnitude, while we can infer that the variance of $\hat\beta$ shrinks with the increase of window size in a fast rate.

\subsubsection*{Negative Binomial regression}
The Negative Binomial regression model assume the dependent variable $y_t$ is Negative Binomial distributed given $x_t$: $y_t | x_t \sim \text{NegBin}(\mu(x_t), r)$. The assumption is the counts are overdispersed - meaning the variance is larger than the expectation. This assumption fits into many real world settings. Comparing with Poisson, the addition of overdispersion parameter $r$ brings more flexibility while also increased computational complexity. Same as Poisson model, Negative Binomial is also able to handle moderate amount of zero counts. The regression model is $y_t \sim \text{NegBin}(\mu_t, r)$ where $\log(\mu_t) = \alpha + x_t\beta, \mu_t = e^{\alpha + x_t\beta}$. $\mathbb{V}(y_t) = \mu_t + \frac{\mu_t^2}{r}$. The score function $U(\beta) = \sum_{i=1}^n (y_i - e^{\alpha + i\beta}) i$. The Fisher information is $I(\begin{bmatrix}
\hat\alpha\\
\hat\beta
\end{bmatrix}) = \sum_{i=1}^n \left(e^{\alpha + i\beta} + \frac{e^{2(\alpha + i\beta)}}{r}\right) \begin{bmatrix}
1 & i \\
i & i^2
\end{bmatrix}$. The variance of $\hat\beta$ also shrinks fast with the increase of window size, in similar way to Poisson Regression, while the detailed formula too complicated to present here.

Define $c = \frac{1}{r}$, for the estimation of $c$, an alternating iteration process is used by the R package MASS \cite{venables2013modern}: For given $c$, the Negative Binomial Regression is then fitted; for fixed means the $c$ is estimated using score and information iterations. Breslow \cite{breslow1984extra} suggested that, given estimation of $\mu(x_t)$, $c$ can be solved by the moment equation $\sum_{t = 1}^{n}\frac{(y_t - \mu(x_t))^2}{\mu(x_t)(1 + c\mu(x_t))} = n - 1$ \cite{nakashima1997some, aeberhard2014robust}. If $x_t$ is a $p$ dimensional vector rather than a scalar, the right hand side should be the degree of freedom as $n - p$. Carroll and Ruppert \cite{carroll1982robust} suggested another unbiased estimating equation originally derived from the score equation of the approximate normal log likelihood, i.e., the pseudo-likelihood equation $\sum_{t = 1}^{n}\frac{(y_t - \mu(x_t))^2 - (1 + c\mu(x_t))}{2(1 + c\mu(x_t))^2} = 0$ \cite{nakashima1997some}.

As the final result is not sensitive to $c$ estimation, we don't need to iteratively optimize over $c$ and $\mu$. Instead, we use an arbitrary plug-in of $c$ based on the previous idea with a rough estimation of $\mu(x_t)$ from Poisson regression, reducing the cost of solving. There are more arbitrary estimations of $c$ for hypothesis testing purposes when the null hypothesis being $c = 0$. For example, \cite{cameron2013regression} suggest a means to calculate $c$ using a technique they call auxiliary Ordinary Least Squares (OLS) regression regression without a constant. If one wants to avoid completely solving Negative Binomial Regression, since $\mathbb{V}(y_t) = \mu(x_t) + c\mu^2_t(x_t)$, the test can still be constructed starting with a Poisson model with fitted $\mu(x_t) = e^{\alpha + x_t\beta}$, and performing auxiliary OLS to estimate $c$ that is to be tested.

$$\frac{(y_t - \mu(x_t))^2 - y_t}{\mu(x_t)} = c\mu(x_t) + \epsilon_t$$ where $\epsilon_t$ is an error term. What's more, \cite{dean1989tests} proposed more constructions for testing. For our purpose, we want to be more honest about the estimation rather than the asymptotic distribution and convergence under null. We propose this formula with derivation. 
\begin{align*}
    \mathbb{E}(y_t|x_t) &= \mu(x_t), \mathbb{V}(y_t|x_t) = \mu(x_t) + c\mu^2_t(x_t)\\
    \mathbb{E}(y_t) &= \mathbb{E}(\mathbb{E}(y_t|x_t))\\
    \mathbb{V}(y_t) &= \mathbb{E}(\mathbb{V}(y_t|x_t)) + \mathbb{V}(\mathbb{E}(y_t|x_t))\\
    &= \mathbb{E}(\mu(x_t) + c\mu^2_t(x_t)) + \mathbb{V}(\mu(x_t))\\
    &= \mathbb{E}(\mu(x_t)) + c\mathbb{E}(\mu^2_t(x_t)) + \mathbb{E}(\mu^2_t(x_t)) - \mathbb{E}^2(\mu(x_t))\\
    \hat c &= \frac{\mathbb{V}(y_t) - \mathbb{E}(\mu(x_t)) + \mathbb{E}^2(\mu(x_t))}{\mathbb{E}(\mu^2_t(x_t))} - 1\\
\end{align*}
\subsubsection*{Poisson smoothing}
Here we assume Poisson distributed counts $y_t \sim \text{Poi}(\mu_t)$, and that the weekday effect is multiplicative. Formally, 
\begin{align*}
\log \mu_t &= \alpha_{\text{wd}(t)} + \log \phi_t
\end{align*}
where $\text{wd}(t) \in \{0, \dots, 6\}$ is the day-of-week of time $t$, $\alpha_{\text{wd}(t)}$ is the corresponding weekday correction, and $\phi_t$ is the latent disease burden at time $t$. 
The Poisson likelihood function $\ell(\alpha,\phi|y)$ for a sample $y = \{y_1, y_2, \ldots, y_T\}$ is $\ell(\alpha,\phi|y) = \prod_{t=1}^T \frac{\mu_t^{y_t} e^{-\mu_t}}{y_t!}$, and $\log \ell (\alpha,\phi|y) = \sum_{t=1}^T(y_t\log \mu_t - \mu_t - \log(y_t!))$. 
For simplicity, we assume that the weekday parameters are identical across time and location.

To fit the $\alpha$ parameters, we minimize the following convex objective function: $f(\alpha, \phi | y) = -\log \ell (\alpha,\phi|y) + \lambda ||\Delta^3 \phi||_1$. $\Delta^3 \phi$ is the third difference of $\phi$. For identifiability, we constrain the sum of $\alpha$ to be zero by setting Sunday’s fixed effect to be the negative sum of the other weekdays. The penalty term $\lambda$ encourages the $\phi$ curve to be smooth and produce meaningful $\alpha$ values.

After estimating $\alpha$, we produce adjusted counts $\log \xi_t = \log y_t - \alpha_{\text{wd}(t)}$, where $\xi_t$ is the output of Delphi’s Epidata API (corrected for day-of-week effects), retaining the Poisson noise. The underlying Poisson rate is $\log \phi_t = \log \mu_t - \alpha_{\text{wd}(t)}$. The adjusted counts $\xi_t$ are inherently smoothed.
\subsubsection*{Log-Normal smoothing}
Real-world data often contains overdispersion, resulting in poor model fit, with unacceptably high deviance values $2 \sum_{t=1}^T \left[y_t \log\left(\frac{y_t}{\mu_t}\right) - (y_t - \mu_t)\right]$ (twice the difference between the log-likelihood of the saturated model and the fitted model) and high Pearson chi-squared statistics $\chi^2 = \sum_{t=1}^T \frac{(y_t - \mu_t)^2}{mu_t}$. Although the Negative Binomial model is popular for addressing overdispersion, the Log-Normal model serves as a good approximation when counts are large, as is the case with most trends of interest.

We assume the counts are generated by a multiplicative noise model $
y_t = \mu_t\cdot\eta_t,\  \log\eta_t \sim N(0, \sigma^2)$, thus, $\log y_t \sim N(\log \mu_t, \sigma^2)$. 
We model the weekday effect as
\begin{align*}
\log y_t &= \log\mu_t + \log\eta_t,\  \log y_t \sim N(\log\mu_t, \sigma^2)\\
\log \mu_t &= \alpha_{\text{wd}(t)} + \log \phi_t
\end{align*}
The Gaussian likelihood function $\ell(\alpha, \phi, \sigma^2 | y) = \prod_{t=1}^T \frac{1}{\sqrt{2\pi \sigma^2}} \exp\left(-\frac{(\log y_t - \log \mu_t)^2}{2\sigma^2}\right)$, and $\log \ell (\alpha, \phi, \sigma^2 | y) = \sum_{t=1}^T \left(-\frac{1}{2\sigma^2} (\log y_t - \log \mu_t)^2 - \log(\sqrt{2\pi \sigma^2})\right)$. 
To estimate $\alpha$ and $\phi_t$, we minimize $f(\alpha, \phi | y, \hat\sigma^2) = -\log \ell (\alpha, \phi, \hat\sigma^2 | y) + \lambda ||\Delta^3 \phi||_1$, where $\hat{\sigma}^2$ is estimated separately and updated iteratively until convergence. We note the negative log-likelihood is convex with respect to $\log \mu_t$ and $\frac{1}{\sigma^2}$ individually but not jointly. We constrain the sum of $\alpha$ to zero the same as in Poisson smoothing.

\subsubsection*{Negative Binomial Smoothing}
Assuming $y_t$ follows a Negative Binomial distribution, suitable for overdispersed count data, a Negative Binomial distribution can be parameterized in different ways, but a common parameterization is $y \sim \text{NegBin}(r, p)$ where $r$ is the number of failures until the experiment is stopped and $p$ is the probability of success. The mean $\mu$ of $y$ in this parameterization is given by $\mu = \frac{r(1-p)}{p}$. Alternatively, it can be reparameterized  by replacing the probability of success $p$ with the mean $\mu$ and keeping the dispersion parameter $r$, which is also called the shape parameter. Larger $r$ indicates less dispersion. In this model, $y_t \sim \text{NegBin}(\mu_t, r)$. 

We model the weekday effect as $\log \mu_t = \alpha_{\text{wd}(t)} + \log \phi_t$, where $\text{wd}(t) \in \{0, \dots, 6\}$ is the day-of-week of time $t$, $\alpha_{\text{wd}(t)}$ is the corresponding weekday correction, and $\phi_t$ is the corrected counts at time $t$. The Negative Binomial likelihood function $\ell(\alpha,\phi|y)$ for a sample $y = \{y_1, y_2, \ldots, y_T\}$ is $\ell(\alpha, \phi, r|y) = \prod_{t=1}^T \binom{y_t + r - 1}{y_t} \left(\frac{r}{r + \mu_t}\right)^r \left(\frac{\mu_t}{r + \mu_t}\right)^{y_t}$, and $$\log \ell (\alpha, \phi, r|y) = \sum_{t=1}^T \left(\log \binom{y_t + r - 1}{y_t} + r \log \left(\frac{r}{r + \mu_t}\right) + y_t \log \left(\frac{\mu_t}{r + \mu_t}\right) \right).$$ 

Estimating the parameter $r$ cannot be done in closed form and requires iterative techniques like Newton's method or the expectation–maximization algorithm. The computational complexity, especially with numerous unknown $\mu_t$ values, makes direct implementation challenging. 

There are also other estimating strategies for joint estimation of the mean and dispersion parameters, like Extended Quasi-Likelihood Estimator proposed by \cite{clark1989estimation},  Extended Double-Extended Quasi-Likelihood Estimator proposed by \cite{lee2001hierarchical}, and Bias-Corrected Maximum Likelihood Estimator proposed by \cite{saha2005bias}. Even with simplifications like factorials replaced by Stirling approximations that adds errors for small counts, the coumputation is still expensive. Even after estimating $r$, solving for $\mu_t$ remains computationally expensive due to the non-convexity of the log-likelihood with respect to $\mu_t$. As a result, while theoretically compelling, Negative Binomial smoothing is left for future work due to its computational intricacies. 

Over-dispersed Poisson is another alternative for handling overdispersion. \cite{kim2021prediction} propose to use the multiplication of a Poisson variable $y^*_t \sim \text{Poi}(\mu_t)$ and a Gamma variable $z = \text{Gam}(c, c)$ to model overdispersion, since $\mathbb{E}(z) = 1$ and variance $\mathbb{V}(z) = \frac{1}{c}$. The distribution is also known as the compound Poisson-Gamma or Poisson-Tweedie distribution. This distribution is a specific case of a Tweedie distribution. Then $y_t = z y^*_t$ has $\mathbb{E}(y_t) = \mathbb{E}(z)\mathbb{E}(y^*_t) = \mu_t$, $\mathbb{V}(y_t) = \mathbb{E}(z^2y^{*2}_t) - \mathbb{E}^2(z y^*_t) = \mathbb{E}(z^2)\mathbb{E}(y^{*2}_t) - \mu_t^2 = \left(\mathbb{V}(z) + \mathbb{E}^2(z)\right)\left(\mathbb{V}(y^*_t) + \mathbb{E}^2(y^*_t)\right) - \mu_t^2 = (\frac{1}{c} + 1)(\mu_t + \mu^2_t) - \mu_t^2 = (\frac{1}{c} + 1)\mu_t + \frac{1}{c}\mu^2_t$. Therefore, the probability mass function is $f(y_t = k) = \int_{0}^{\infty} f_{\text{Poi}}(k; \mu_t z) f_{\text{Gam}}(z; c, c) \, dz$, where $f_{\text{Poi}}(k; \lambda) = e^{-\lambda} \frac{\lambda^k}{k!}$ is the PMF of the Poisson distribution, $f_{\text{Gam}}(z; c, c) = \frac{c^c}{\Gamma(c)} z^{c-1} e^{-cz}$ is the PDF of the Gamma distribution. However, this formula is even more complicated than Negative Binomial, thus we leave it to future work.

\end{document}